\documentclass{article}

\PassOptionsToPackage{numbers, compress}{natbib}
\usepackage[final]{nips_2016_mine}

\usepackage[utf8]{inputenc} % allow utf-8 input
\usepackage[T1]{fontenc}    % use 8-bit T1 fonts
\usepackage{hyperref}       % hyperlinks
\usepackage{url}            % simple URL typesetting
\usepackage{booktabs}       % professional-quality tables
\usepackage{amsfonts}       % blackboard math symbols
\usepackage{nicefrac}       % compact symbols for 1/2, etc.
\usepackage{microtype}      % microtypography

\usepackage{setspace}

\usepackage{qiangstyle}

\title{Particle Variational Inference Via Measure Transport: A New Framework}

\title{A Population Gradient Descent for Variational Inference}

\title{Scalable Bayesian Variational Inference Using Transport Gradient Descent}

\title{Bayesian Variational Inference is as Easy as Gradient Descent}

\title{Stein Variational Gradient Descent: A General Purpose Bayesian Inference Algorithm}

\author{
Qiang Liu ~~~~~~~~~ Dilin Wang\\
  Department of Computer Science\\
  Dartmouth College\\
  Hanover, NH  03755\\
  \texttt{\{qiang.liu,  dilin.wang.gr\}@dartmouth.edu} \\
}

\begin{document}
% \nipsfinalcopy is no longer used

\maketitle

\begin{abstract}
We propose a general purpose variational inference algorithm that forms a natural counterpart of gradient descent for optimization. 
% based on evolving a set of particles to minimize KL divergence. %iterative transport a set of particles to the target distribution. 
Our method iteratively transports a set of particles to match the target distribution, 
by applying a form of functional gradient descent that minimizes the KL divergence. % of the true and the particle distributions. 
%between the true posterior and the distribution that the particle represents. 
%Our method uses a set of particles for approximation, on which incremental transforms (in the form of a gradient descent) are applied recursively to mininimize the KL divergence between the true posterior and the distribution that the particle represents. 
%Instead of assuming a parametric form on the variational reference distribution, we recursively apply transforms on a set of particles  to minimize the KL divergence in the fastest direction. 
Empirical studies are performed on various real world models and datasets, on which our method is competitive with existing state-of-the-art methods. 
%comparisons with existing methods are performed on various real world models, including a Bayesian neural network on which we are competitive with state-of-the-art methods. 
The derivation of our method is based on a new theoretical result %generation of \emph{de Bruijn identity} 
that connects the derivative of KL divergence under smooth transforms with Stein's identity and a recently proposed kernelized Stein discrepancy, which is of independent interest. 
%KL divergence with a recently proposed kernelized Stein discrepancy, which is of independent interest. 
%
%
%that connects KL divergence with a recently proposed kernelized Stein discrepancy, which is of independent interest. 
%this is in contract with the traditional variational inference that based on a predefined parametric form. 
%Our key result is based on a generation of deBruijn identity that connects the derivative of KL divergence under smooth transform with Stein's identity. 
%Throughout theoretical analysis and experiments are provided. 
\end{abstract}

\section{Introduction}

Bayesian inference provides a powerful tool for modeling complex data and reasoning under uncertainty, 
but casts a long standing challenge on computing intractable posterior distributions. 
%The widely used Markov chain Monte Carlo (MCMC) methods draw approximate samples by simulating Markov chains,  
Markov chain Monte Carlo (MCMC) has been widely used to draw approximate posterior samples,  
%The Markov chain Monte Carlo (MCMC) methods have been the default tool for drawing approximate posterior samples, %by simulating Markov chains, 
%draw samples by simulating Markov chains, but 
but is often slow and has difficulty accessing the convergence. 
%The variational inference methods provide an determinstic alternative by framing instead frame the Bayesian inference problem into an deterministic optimization problem that involves approximating the target distribution with a simpler distributions by minimizing their KL divergence. 
Variational inference instead frames the Bayesian inference problem into a deterministic optimization that approximates the target distribution with a simpler distribution by minimizing their KL divergence. 
This makes variational methods efficiently solvable by using off-the-shelf optimization techniques, %for which convergence can be easily accessed. 
%In addition, 
%variational inference 
and easily applicable to large datasets (i.e., "big data") using the stochastic gradient descent trick \citep[e.g.,][]{hoffman2013stochastic}. In contrast, it is much more challenging to scale up MCMC to big data settings \citep[see e.g.,][]{welling2011bayesian, firefly}. 
%Meanwhile, variational methods critically depends on the 
%reviews: jun zhu big data big learning; monte carlo for tall data; Adam, patterns in Monte Carlo big data}. 

Meanwhile, both the accuracy and computational cost of variational inference critically depend on the set of distributions in which the approximation 
is defined. Simple approximation sets, such as these used in the traditional mean field methods, are too restrictive to resemble the true posterior distributions, 
while more advanced choices cast more difficulties on the subsequent optimization tasks. 
%such as these based on mixture or hierarchical distributions \citep[e.g.,][]{jaakkola1999improving, gershman2012nonparametric, ranganath2016hierarchical}, 
%copula models \citep[e.g.,][]{kingma2013auto, jimenez2015variational, mnih2014neural}, 
%such as these based on mixture distributions \citep[e.g.,][]{jaakkola1999improving, gershman2012nonparametric} or neural networks \citep[e.g.,][]{kingma2013auto, jimenez2015variational, mnih2014neural}, 
%cast more difficulties on the subsequent optimization tasks. 
%
\todo{This makes it difficult for the algorithm designers to strike the right balance, and efficient variational methods are often need to be derived in a model by model basis.}
For this reason, efficient variational methods often need to be derived on a model-by-model basis, 
 %, and more critically, 
causing is a major barrier for developing general purpose, user-friendly variational tools 
applicable for different kinds of models, and 
accessible to non-ML experts in application domains. %(continuous variable 
%There is a significant recent interest on developing user-friendly variational methods
%A line of recent advances has been made to 
%It is useful 

% salimans2015markov Salimans, MCMC and variational, bridging the gap. http://jmlr.org/proceedings/papers/v37/salimans15.pdf
%ranganath2016hierarchical: use hierarchical models as propsoal. 
%
%
%
%

This case is in contrast with the maximum \emph{a posteriori} (MAP) optimization tasks for finding the posterior mode (sometimes known as the \emph{poor man's Bayesian estimator}, in contrast with the \emph{full Bayesian inference} for approximating the full posterior distribution), 
%(sometimes known as the \emph{poor man's Bayesian estimator}, which only finds the mode, instead of approximating the full distribution), 
for which variants of (stochastic) gradient descent serve as a simple, generic, yet extremely powerful toolbox. 
There has been a recent growth of interest in creating user-friendly variational inference tools \citep[e.g.,][]{ranganath2013black, gershman2012nonparametric, kucukelbir2015automatic, dai2016provable}, 
but more efforts are still needed to develop more efficient general purpose algorithms. %that works in most cases. 
%efficient and general purpose algorithms, mostly variants of (stochastic) gradient descent exists. 
%
%for most distributions. 
%needed to create general purpose tools for arbitrary continuous distributions. 
\todo{Similar efforts have made for general purpose MCMC tools \citep{foreman2013emcee, homan2014no},  but they suffer the weakness of MCMC, and have difficulty in scaling to large datasets. }
\todo{
Another challenge with Bayesian inference is to develop general purpose, user-friendly tools accessible to non-experts (continuous variable models). 
%Additionally, it remains a major challenge to develop widely applicable, user-friendly Bayesian inference tools for practitioners in various scientific domains. 
%This has especially been a challenging for variational methods since it often requires expert knowledge to develop variational methods on a model-by-model basis. 
%This has been well
This has been a somewhat easier task for MCMC since the Metropolis-Hasting framework 
is itself generally applicable, and several well packaged tools exist \citet{foreman2013emcee, homan2014no}. 
On the other hand, variational inference requires more model specific considerations due to the need for defining the approximating families and deriving the corresponding optimization algorithms. 
There is a significant recent interest on developing user-friendly variational methods \citep[e.g.,][]{ranganath2013black, gershman2012nonparametric, kucukelbir2015automatic}, but still does not eliminate the need for specifying approximation families. 
}
%Deriving 
%Some general purpose MCMC tools has been exist \citet{foreman2013emcee, homan2014no}, 
%This has been particularly the case for variational inference, which often need to be derived on a model-by-model basis. 
%because of the need for specifying model-specific approximating families and developing the corresponding optimization algorithms. 
%Developing \emph{black-box} variational inference has been a recent focus of research \citep[e.g.,][]{ranganath2013black, gershman2012nonparametric, kucukelbir2015automatic}, with the goal of allowing practitioners to quickly apply the inference tools and test and revise their models. %variational tools. 
%Deriving these algorithms on a model-by-model basis is tedious work. This hinders us from rapidly exploring modeling assumptions when solving applied problems, and it makes variational methods on complicated dis- tributions impractical for many practitioners. Our goal in this paper is to develop a ?black box? variational inference algorithm, a method that can be quickly ap- plied to almost any model and with little effort. Our method allows practitioners to quickly design, apply, and revise models of their data, without painstaking derivations each time they want to adjust the model.

%It is useful to draw analogue to the optimization tasks, for which efficient and general purpose algorithms, mostly variants of (stochastic) gradient descent exists. 
%(in Bayesian setting, this corresponds to maximum a posteriori)
In this work, we propose a new general purpose variational inference algorithm which can be treated as a natural counterpart of gradient descent for full Bayesian inference (see Algorithm \ref{alg:alg1}). Our algorithm uses a set of particles for approximation, on which a form of (functional) gradient descent is performed to minimize the KL divergence and drive the particles to fit the true posterior distribution. 
Our algorithm has a simple form, and can be applied whenever gradient descent can be applied. 
In fact, it reduces to gradient descent for MAP when using only a single particle, while automatically turns into a full Bayesian approach with more particles. 

\todo{In this work,  we propose a new variational inference algorithm that is based on minimizing the KL divergence of a population of particles and the target distribution. 
We maintain a set of particles, and apply transforms sequentially to minimize KL divergence. We find a closed form solution for the optimal local transform that gives the steepest descent on the KL divergence in a kernel space. This yields a simple algorithm with a (kernelized) gradient descent. Interestingly, our algorithm automatically reduces to the typical gradient descent for penalized maximum likelihood (i.e., finding the posterior mode) with one particle (or with a kernel of zero bandwidth). This allows us provides a practical computational procedure that smooths between penalized maximum likelihood and Bayesian methods by simply adding the number of particles. Our algorithm is therefore well suited for automatic implementation. }
Underlying our algorithm is 
a new theoretical result 
%a new generation of \emph{de Bruijn identity}
that connects the derivative of KL divergence w.r.t. smooth variable transforms and a recently introduced kernelized Stein discrepancy \citep{liu2016kernelized, oates2014control,  chwialkowski2016kernel}, 
which allows us to derive a closed form solution for the optimal smooth perturbation direction 
that gives the steepest descent on the KL divergence
within the unit ball of a reproducing kernel Hilbert space (RKHS). 
%within a 
%functional gradient to give the steepest descent on the KL divergence within a 
% in a kernel space. 
%which allows us to derive a closed form solution for the optimal local transform within a reproducing kernel Hilbert space (RKHS) that defines the steepest descent on the KL divergence in a kernel space. 
This new result is of independent interest, and can find wide application in machine learning and statistics beyond variational inference. 
%This close form makes our algorithm highly efficient, and ease to implement. 
%We expect that this result will find more general application beyond variational inference. 

%*The idea of variational inference by recursively transforming particles has been discussed in several recent work \citep{jimenez2015variational,  marzouk2016introduction}. *Our key contribution is to reveal a theoretical connection between  the derivatives of KL divergence w.r.t. smooth transforms and a recently introduced kernelized Stein discrepancy \cite{oates2014control,  chwialkowski2016kernel, liu2016kernelized}; this result extends the well known connection of derivative of KL divergence and Fisher divergence, and deBurijn identity. Using the result in \citet{liu2016kernelized}, we can find the a closed form solution for the optimal local transform that gives the steepest descent on the KL divergence in a kernel space. This close form makes our algorithm highly efficient, and ease to implement. We expect that this result will find more general application beyond variational inference. 

\paragraph{Outline}
This paper is organized as follows. 
Section~\ref{sec:background} introduces backgrounds on kernelized Stein discrepancy (KSD). 
Our main results are presented in Section~\ref{sec:method} in which we clarify the connection between KSD and KL divergence, and leverage it to develop our novel variational inference method. 
Section~\ref{sec:related} discusses related works, and Section~\ref{sec:experiments} presents numerical results. 
The paper is concluded in Section~\ref{sec:conclusion}. 
%The conclusion in drawn in Section~\ref{sec:conclusion}. 

\section{Background}
\label{sec:background}
%We give a brief introduction on Stein's identity and the recently proposed kernelized Stein discrepancy (KSD) in Section~\ref{sec:stein}, and then  variational inference in Section~\ref{sec:variational}. 
\paragraph{Preliminary}
Let $x$ be a continuous random variable or parameter of interest taking values in $\X \subset \R^d$, and $\{D_k\}$ is a set of i.i.d. observation. 
With prior $p_0(x)$, Bayesian inference of $x$ involves reasoning with the posterior distribution $p(x) \coloneqq \bar p(x) / Z$ with $\bar p (x) \coloneqq  p_0(\x) \prod_{k=1}^N p(D_k  | \x)$, 
where $Z = \int \bar p(x) dx$ is the troublesome normalization constant. %whose computation is often critically challenging.
We have dropped the conditioning on data $\{D_k\}$ in $p(x)$ for convenience. 

Let $\k(x,x') \colon \X \times \X \to \R$ be a positive definite kernel. 
The reproducing kernel Hilbert space (RKHS) $\H$ of $\k(x,x')$ is the closure of linear span $\{f \colon  f(x)= \sum_{i=1}^m a_i \k(x, x_i), ~~~ a_i \in \R, ~~ m\in \mathbb{N}, ~ x_i \in \X  \}$, equipped with inner products $\la f, ~g \ra_{\H}= \sum_{ij} a_i b_j \k(x_i,x_j)$ for $g(x)  = \sum_i b_i \k(x,x_i)$. 
Denote by $\H^d$ the space of vector functions $\vv f = [f_1,\ldots, f_d]$ with $f_i \in \H$, equipped with inner product $\la \vv f, \vv g \ra_{\H^d} = \sum_{i=1}^d \la f_i, g_i \ra_\H$.  
We assume all the vectors are column vectors. 
%We first introduce Stein's identity and kernelized Stein discrepancy (KSD), and then give a brief introduction on Bayesian variational inference.
% in Section~\ref{sec:variational}. 
% \citep{liu2016kernelized, oates2014control, chwialkowski2016kernel}. 
%In this work, we will clarify the connection between KSD and KL divergence, and leverage it to develop our novel variational inference method. 

%We are interested in estimating 
%Our method is based on the following observation. 

%\paragraph{Notation}
%\subsection{Kernelized Stein Discrepancy}
\paragraph{Stein's Identity and Kernelized Stein Discrepancy}
\label{sec:stein}
\myempty{
Let $\k(x,x') \colon \X \times \X \to \R$ be a positive definite kernel; we denote by $\k(x, \cdot)$ the one-variable function with fixed $x$. 
The reproducing kernel Hilbert space (RKHS) $\H$ of $\k(x,x')$ is the closure of linear span $\{f \colon  f= \sum_{i=1}^m a_i \k(x, x_i), ~~~ a_i \in \R, ~~ m\in \mathbb{N}, ~ x_i \in \X  \}$, equipped with inner products $\la f, ~g \ra_{\H}= \sum_{ij} a_i b_j \k(x_i,x_j)$ for $g  = \sum_i b_i \k(x,x_i)$. One can verify that such $\H$ has a \emph{reproducing} property in that $f = \la f, ~ \k(x, \cdot)\ra_{\H}$. 
}

%We give a brief introduction on Stein's identity and the recently proposed kernelized Stein discrepancy (KSD) \citep{liu2016kernelized, oates2014control, chwialkowski2016kernel}. In this work, we will clarify the connection between KSD and KL divergence, and leverage it to develop our novel variational inference method. 
% that forms the foundation of our method. 
%We introduce kernelized Stein discrepancy (KSD) that plays a key role in our method. 
Stein's identity plays a fundamental role in our framework. 
Let $p(x)$ be a continuously differentiable (also called smooth) density supported on $\X\subseteq \RR^d$, and 
$\ff(x)=[\f_1(x), \cdots, \f_d(x)]^\top$ a smooth vector function. Stein's identity states that for sufficiently regular $\ff$, we have 
\begin{align}\label{equ:steq1}
\E_{x\sim p}[\stein_p \ff(x)] = 0, && \text{where} && 
\stein_p \ff(x) = \ff(x)  \nabla_x \log p(x)^\top  + \nabla_x \ff(x), 
\end{align}
where $\stein_p$ is called the Stein operator, which acts on function $\ff$ and yields a zero mean function $\stein_p \ff(x) $ under $x\sim p$.  
This identity can be easily checked using integration by parts, assuming mild zero boundary conditions on $\ff$, either $p(x)\ff(x) = 0$, $\forall x \in \partial \X$ when $\X$ is compact, or $\lim_{||x||\to \infty} \ff(x)p(x) = 0 $ when $\X = \R^d$. 
%$\lim_{||x|| \to \infty} \ff(x) p(x) = 0$ on $\ff$. 
We call  that $\ff$ is in the Stein class of $p$ if Stein's identity \eqref{equ:steq1} holds. 

Now let $q(x)$ be a different smooth density also supported in $\X$, and 
consider the expectation of $\stein_p \ff(x)$ under $x\sim q$, then $\E_{x\sim q} [\stein_p \ff(x)]$ would no longer equal zero for general $\ff$. 
Instead, the magnitude of $\E_{x\sim q} [\stein_p \ff(x)]$ relates to how different $p$ and $q$ are, and can be leveraged to define a discrepancy measure, known as \emph{Stein discrepancy}, %between $p$ and $q$, 
by 
%Stein discrepancy has a general form of $\max_{\ff  \in \F}\E_{x\sim q} [\stein_p \ff(x)]$, 
considering the ``maximum violation of Stein's identity'' for $\ff$ in some proper function set $\F$:
\begin{align*}
%\label{equ:ksdexp}
{\S(q, ~p)} = \max_{\ff \in \F } \big \{ [\E_{x\sim q}\trace(\stein_p \ff(x))]^2 \big\},  
%\\& \text{where} ~~~~ \stein_p f(x) = \E_{x\sim q}[\score_p(x) f(x)  + \nabla_x f(x)], 
\end{align*}
Here the choice of this function set $\F$ is critical, and decides the discriminative power and computational tractability of Stein discrepancy. 
Traditionally, $\F$ is taken to be sets of functions with bounded Lipschitz norms, which unfortunately casts a challenging functional optimization problem that is computationally intractable or requires special considerations (see \citet{gorham2015measuring} and reference therein). 
%where the expectation in taking under $q(x)$
%When $p\neq q$ \eqref{equ:steq1} implies that there  exists some $\ff$ such that $\E_{x\sim q} [\stein_p \ff(x)] \neq 0$. This allows us to define a discrepancy measure, known as \emph{Stein discrepancy}, by considering the maximum violation of Stein identity $\max_{\ff  \in \F}\E_{x\sim q} [\stein_p \ff(x)] $ in some proper function set $\F$. 
%Traditional definitions of Stein discrepancy consider sets of functions with bounded Lipschitz norms, and cast challenging functional optimization problems (see \cite{gorham2015measuring} and reference therein). 

Kernelized Stein discrepancy bypasses this difficulty by maximizing $\ff$ in the unit ball of a reproducing kernel Hilbert space (RKHS) for which the optimization has a closed form solution. Following \citet{liu2016kernelized}, KSD is defined as %Let H be a RKHS associated with a positive definite kernel
\begin{align}
\label{equ:ksdexp}
& {\S(q, ~p)} = \max_{\ff \in \H^d } \big \{ [\E_{x\sim q}(\trace(\stein_p \ff(x)))]^2 , ~~~~~~s.t. ~~~~~~ ||\ff ||_{\H^d} \leq 1 \big\},  
%\\& \text{where} ~~~~ \stein_p f(x) = \E_{x\sim q}[\score_p(x) f(x)  + \nabla_x f(x)], 
\end{align}
where we assume the kernel $k(x,x')$ of RKHS $\H$ is 
%where  $\H$ is the RKHS of kernel $k(x,x')$, 
%%for which we assume each $k(x, x')$ 
%which is assumed to be 
in the Stein class of $p$ as a function of $x$ for any fixed $x'\in \X$. 
The optimal solution of \eqref{equ:ksdexp} has been shown \citep{liu2016kernelized, oates2014control, chwialkowski2016kernel} to be  $\ff(x)= \ff^*_{q,p}(x) /  ||\ff^*_{q,p}||_{\H^d}$, where
%It has been showed that the above optimization in fact has a closed form solution \citep[e.g., Theorem 3.7 of][]{liu2016kernelized}  $ \ff^*(x) /  ||\ff^*||_{\H^d}$, where
\begin{align}\label{equ:phiqp00}
\ff ^*_{q,p}(\cdot)= 
\E_{x \sim q} [\stein_p k(x,\cdot)], 
%= \E_{x \sim q}[\nabla_{x}\log p(x) k(x, \cdot)  + \nabla_{x} k(x, \cdot)],  
&&
\text{for which we have}
&&
\S(q, ~ p) = || \ff^*_{q,p} ||_{\H^d}^2. 
%f^*(x)  = \frac{1}{ ||\ff^*||_{\H^d}} \ff^*(x) , &&
%\text{where} &&
\end{align}
%which correspondingly also gives a closed form for $\S(q, ~ p)$ \citep[see e.g., Theorem 3.7 of][]{liu2016kernelized}. 
%We will leverage this
%$\S(q,~p)$ is always nonnegative because $\kp(x,x')$ is positive semi-definite if $\k(x,x')$ is positive semi-definite \citep[e.g.,][Theorem 3.6]{liu2016kernelized}.
One can further show that $\S(q,~p)$ equals zero (and equivalently $\ff^*_{q,p}(x)\equiv0$) if and only if $p = q$ once $\k(x,x')$ is strictly positive definite in a proper sense \citep[See][]{liu2016kernelized, chwialkowski2016kernel}, 
which is satisfied by commonly used kernels such as 
 the RBF kernel $\k(x, x') = \exp(-\frac{1}{h} || x  - x'||^2_2)$. Note that the RBF kernel is also in the Stein class of smooth densities supported in $\X = \R^d$ because of its decaying property. 
%strictly integrally positive definite in \citet{liu2016kernelized} and $cc$-universal in \citet{chwialkowski2016kernel}; 
%\citet{liu2016kernelized} requires $\k(x,x')$ to be strictly integrally positive definite and \citet{chwialkowski2016kernel} requires $k(x,x')$ to be $cc$-universal; 
% these conditions are satisfied by the RBF kernel $\k(x, x') = \exp(-\frac{1}{h} || x  - x'||^2_2)$, which is also in the Stein class of smooth densities supported in $\X = \R^d$ because of its decaying property. 

Both Stein operator and KSD depend on $p$ only through the score function $\nabla_x \log p(x)$, which can be calculated without knowing the normalization constant of $p$, 
because we have $\nabla_x \log p(x) = \nabla_x \log \bar p(x)$ when $p(x) = \bar p(x)/Z$. 
%which does not depend on the normalization constant in $p(x)$, that is, when $p(x) = \bar p(x)/Z$ it equals $\nabla_x \log \bar  p(x)$, independent of the normalization constant $Z$ that is often critically difficult to calculate. 
This property makes Stein's identity a powerful tool for handling unnormalized distributions that appear widely in machine learning and statistics. 
%One important feature of Stein operator and KSD is that calculating $\nabla_x \log p (x)$ does not depend on the normalization constant in $p(x)$, that is, when $p(x) = \bar p(x)/Z$ it equals $\nabla_x \log \bar  p(x)$, independent of the normalization constant $Z$ that is often critically difficult to calculate. This property makes Stein's identity a powerful tool for handling unnormalized distributions that appear widely in machine learning and statistics. 

%\paragraph{Bayesian Variational Inference}
%\subsection{Bayesian Variational Inference}
%\label{sec:variational}
%Let $x$ be a variable or parameter of interest with prior $p_0(x)$, and $\{D_k\}$ is a set of i.i.d. observation. Bayesian inference of $x$ involves reasoning with the posterior distribution $p(x) \coloneqq \bar p(x) / Z$ with $\bar p (x) \coloneqq  p_0(\x) \prod_{k=1}^N p(D_k  | \x)$, where $Z = \int \bar p(x) dx$ is the troublesome normalization constant. %whose computation is often critically challenging. We have dropped the conditioning on data $\{D_k\}$ in $p(x)$ for convenience. 
%Let $p(\x) = \tilde p(x) / Z$ be a probability density of interest supported on open set $\X \subset \RR^d$, and we are interested in evaluating expectations of form $\E_p h = \int h(\x) p(\x) d\x$;
%here $Z$ is the normalization constant $Z = \int f(x) dx$ whose computation is often critically challenging. 
%For Bayesian inference problems, we often have $f(\x) \coloneqq  p_0(\x) \prod_{k=1}^N p(D_i  | \x)$ where $p_0(\x)$ is a prior distribution, and $\{D_i\}_{i=1}^N$  is a set of observed data. 

\section{Variational Inference Using Smooth Transforms}
\label{sec:method}
Variational inference approximates the target distribution $p(x)$ using a simpler distribution $q^*(x)$ found in a predefined set $\mathcal Q = \{q(x)  \}$
of distributions by minimizing the KL divergence, that is, 
%a simpler distribution $q(x)$ by minimizing their KL divergence. 
%Let $\mathcal Q = \{q(x)  \}$ be a set of simple distributions, % indexed by parameter $\para$, 
%variational inference solves the following optimization problem: 
\begin{align}\label{equ:kl}
q^* = \argmin_{q \in \mathcal Q} \big\{ \KL(q ~||~ p) \equiv   \E_q[\log q(x)]  - \E_q [\log \bar p(x)]  + \log Z \big\}, 
\end{align}
where we do not need to calculate the constant $\log Z$ for solving the optimization. 
%we dropped the troublesome normalization constant $Z$ because it is not involved in the optimization. 
%s
The choice of set $\mathcal Q$ is critical and defines different types of variational inference methods. %(see Section~\ref{sec:related} for a review).  %different variational inference methods. 
%this includes the traditional mean-field, and Gaussian-type variational methods for which $\Q$ consists of fully factorized, or Gaussian (mixture) based distributions \citep[see e.g.,][]{jordan1998learning}, as well as the more recent neural variational inference which defines richer sets $\mathcal Q$ by leveraging the power of neural network architectures \citep[e.g.,][]{kingma2013auto, jimenez2015variational, mnih2014neural}.   
%Ideally, the best set $\mathcal Q$ should be 
The best set $\mathcal Q$ should strike a balance between 
i) \emph{accuracy}, broad enough to closely approximate a large class of target distributions,  %(ideally arbitrary) target distributions, and
ii) \emph{tractability}, consisting of simple distributions that are easy for inference, and 
iii) \emph{solvability} so that the subsequent KL minimization problem can be efficiently solved.  
%Ideally, the best set $\mathcal Q$ should be 
%i) \emph{tractable}, consisting of simple distributions that are easy for inference, 
%ii) \emph{universal}, broad enough to form close approximation for a large class of target distribution, and %(ideally arbitrary) target distributions, and
%iii) \emph{solvable}, so that the subsequent KL minimization problem can be efficiently solved.  
%strike a balance between the flexibility for close approximation of the target distribution as well as the computational tractability for the followed KL minimization problem. 
%The choice of the set $\mathcal Q$ of easy distributions forms a key component of the design of variational inference, which should strike a balance between the flexibility for close approximation of the target distribution as well as the computational tractability for the followed KL minimization problem. 
%The design of variational inference then consists of  selecting the easy distributions $\mathcal Q$ to closely approximate $p(x)$, and subsequently solve the resulting variational optimization which often involve further approximation techniques. 
%We provide a comprehensive review of recent advances of variational inference in Section~\ref{sec:related}.

%
%
In this work, we focus on the sets $\Q$ consisting of distributions obtained by smooth transforms %from change of variables 
from a tractable reference distribution, % as recently discussed in \citet{jimenez2015variational}, %as recently studied in \citet{marzouk2016introduction},
that is, we take $\Q$ to be the set of distributions of random variables of form $z = \T(x)$ where $\T\colon \X \to \X$ is a smooth one-to-one transform, and $x$ is drawn from a tractable reference distribution $q_0(x)$.  
%Arguably, such $\Q$ satisfies the following desired properties:
%\emph{Tractability}. By the change of variables formula, the density of $z$ is 
 By the change of variables formula, the density of $z$ is % calculated by
$$
q_{[\T]}( \z ) = q(\T^{-1}(\z)) \cdot |\det(\nabla_\z \T^{-1}(\z))|, 
$$
where $\T^{-1}$ denotes the inverse map of $\T$ and $\nabla_\z \T^{-1}$ the Jacobian matrix of $\T^{-1}$. 
%This defines a set $Q$ of tractable distributions,
%The distributions in such $\Q$ are considered tractable, 
Such distributions are computationally tractable, 
 in the sense that the expectation under $q_{[\T]}$ can be easily evaluated by averaging $\{z_i\}$ when $z_i = \T(x_i)$ and $x_i \sim q_0.$
%Since set of 
%\blue{In addition,  expectations under $q_{[\T]}$ can be calculated even without evaluating the inverse map and the Jacobian: $\E_{z\sim q_{[\T]}}[h(x)] = \E_{x \sim q_0} [h(\T(x))]$ for any $h(\cdot)$; this is known as \emph{the law of unconscious statistician}. }
%\emph{Universality}. 
Such $\Q$ can also in principle closely approximate almost arbitrary distributions: it can be shown that there always exists a measurable transform $\T$ between any two distributions without atoms (i.e. no single point carries a positive mass); in addition, for Lipschitz continuous densities $p$ and $q$, there always exist transforms between them that are least as smooth as both $p$ and $q$. 
We refer the readers to \citet[][]{villani2008optimal} for in-depth discussion on this topic. 
%we refer the readers to \citep[see e.g.,][]{villani2008optimal} for in-depth treatment on the topic. 

%\emph{Solvability}. 
%\change{change}
%\red{It remains a critical challenge to solve the variational optimization in \eqref{equ:kl} to find the optimal transform $\T$.} 
In practice, however, we need to restrict the set of transforms $\T$ properly to make the corresponding variational optimization in \eqref{equ:kl}
practically solvable.  
%in practice. %It remains a critical challenge to define the set of $\T$ properly to allow us to solve the variational optimization in \eqref{equ:kl} in practice. %
One approach is to consider $\T$ with certain parametric form and optimize the corresponding parameters \citep[e.g.,][]{jimenez2015variational, marzouk2016introduction}. 
However, this introduces a difficult problem on selecting the proper parametric family to balance the \blue{accuracy, tractability and solvability}, 
especially considering that $\T$ has to be an one-to-one map and has to have an efficiently computable Jacobian matrix. %%and this is 

%In addition, the requirement that $\T(x)$ has to be one-to-one map casts a restriction on the family one can possibly choose. 
%which however, makes the algorithm complex, and 
%construct such an transport map that transports the mass of $q_0$ to the target distribution.
%One approach was recently studied in \citet{jimenez2015variational}which restricts on transforms of $\T(x)=f_\ell(\cdots (f_1(x)))$, where each $f_i$ is a simple transform with a parametric form. In this way, $\T(x)$ is effectively a multiple layer feedforward network,  

%In this work
Instead, we propose a new algorithm that iteratively constructs incremental transforms that effectively perform steepest descent on $\T$ in RKHS. 
Our algorithm does not require to explicitly specify parametric forms, nor to calculate the Jacobian matrix, and 
has a particularly simple form that mimics the typical gradient descent algorithm, making it easily implementable even for non-experts in variational inference. 
%Our algorithm (shown in Algorithm~\ref{alg:alg1}) has a particular simple form that mimics the simple gradient descent algorithm, and can be easily implemented even for non-experts. 
%; this eliminates the need for assuming parametric forms and  
%Our algorithm has a very simple form of gradient descent, and can be easily implemented by non-experts. 
%It also does not requires to calculate the Jacobian. 
%that assumes  $\T(x)$, by peforming 
%We show that this problem can be solved efficiently using a simple particle gradient descent algorithm that can be easily implemented given $\nabla_x \log p(x)$ as input.  Our method applies incremental transforms on $x$ recursively, where each transform is a simple gradient descent on the particles.  [todo] In our practical algorithm, we maintain a set of particles that is initially drawn from $q_0(x)$, 
% between $q_0$ and 
%include all smooth densities since 
%both of which can be calculated tractably given $q_0$ and $\T(\cdot)$. 
%&&\E_{z\sim q_{[\T]}}[h(x)] = \E_{x \sim q_0} [h(\T(x))],
%which can be calculated tractably, in addition, the expectation under $q_{[\T]}(\z)$ can be simply 
%To be specific, let $q_0(x)$ be a simple reference distribution such as Gaussian or uniform distribution. 
% $\Q$ 

\subsection{Stein Operator as the Derivative of KL Divergence}\label{sec:steinKL}
%We further assume $q(\x)$ to be a computationally tractable distribution supported on $\X$ such as Gaussian or uniform distribution. 
To explain how we 
%develop the transform $T$ to
minimize the KL divergence in \eqref{equ:kl},
we consider an incremental transform formed by a small perturbation of the identity map: $\T (x) = x + \para \ff(x)$, where $\ff(x)$ is a smooth function that characterizes the perturbation direction 
and the scalar $\para$ represents the perturbation magnitude. 
When $|\para|$ is sufficiently small, the Jacobian of $\T$ is full rank (close to the identity matrix), and hence 
$\T$ is guaranteed to be an one-to-one map by the inverse function theorem. 
%since the Jacobian of $\T$ is full rank (close to the identity matrix) when $\para$ is close to zero. 
%Let $q_{[\T]}(z)$ be the density of $\z = T_\para(\x)$ when $x\sim q_0(x)$, that is, $q_{[\T]}( \z ) = q(T^{-1}(\z)) \cdot |\det(\nabla_\z T^{-1}(\z))|. $
%Applying $T_\para$ on $\x$ with density $q(\x)$, then the density of $\z = T_\para(\x)$ is given by 
%We define a transform $$
%We define a transform $T  \colon \X \to \X $ via $T_{\para}(x) = x + \para \ff(x)$. where $\ff \colon  \X \to \X $ is a continuously differentiable function and $\para$ is scalar parameter that is sufficiently small such that $T_{\para}$ is an one-to-one map as implied by the implicit function theorem. Applying $T_\para$ on $\x$ with density $q(\x)$, then the density of $\z = T_\para(\x)$ is given by 
%$$q_{\para}( \z ) = q(T^{-1}_\para(\z)) \cdot |\det(\nabla_\z T^{-1}_\para(\z))|, $$

The following result, which forms the foundation of our method, draws an insightful connection between Stein operator and the derivative of KL divergence w.r.t. the perturbation magnitude $\para$. 
%plays a central role in our method: 
\begin{thm}\label{thm:dkl}
Let $\T(x) = x+ \para \ff(x)$ and $q_{[\T]}(z)$ the density of $z = \T(x)$ when $x\sim q(x)$, we have 
\begin{align}
\label{equ:pdir}
\nabla_{\para}\KL(q_{[\T]} ~||~ p ) ~\big|_{\para = 0} =  - \E_{x\sim q} [\trace(\stein_p \ff(x))], 
\end{align}
where $\stein_p \ff(x) = \nabla_x \log p(x) \ff(x)^\top + \nabla_{x} \ff(x)$ is the Stein operator. 
\end{thm}
%\begin{proof}See Appendix. \end{proof}
%
Relating this to the definition of KSD in \eqref{equ:ksdexp}, 
we can identify the $\ff^*_{q,p}$ in \eqref{equ:phiqp00} as the optimal perturbation direction that gives the steepest descent on the KL divergence %(minimizing \eqref{equ:pdir}) 
in zero-centered balls of $\H^d$. 
%in the unit ball of $\H^d$. % at the fastest speed. 
%and the result of kernelized Stein discrepancy in \citet{liu2016kernelized}, % \eqref{ksdexp}, 
%we can identify the optimal perturbation direction that gives the steepest descent on the KL divergence in the unit ball of a RKHS. % at the fastest speed. 
%The following result is a direct consequence of Theorem~\ref{thm:dkl} and Theorem 3.7 of \citet{liu2016kernelized}.
\begin{lem}\label{lem:dd}
Assume the conditions in Theorem~\ref{thm:dkl}. 
Consider all the perturbation directions $\ff$ in the ball $\B = \{\ff \in \H^d \colon ||\ff ||_{\H^d}^2 \leq \S(q, ~p)\}$ of vector-valued RKHS $\H^d$, the direction of steepest descent that maximizes the 
negative gradient in \eqref{equ:pdir} is the $\ff^*_{q,p}$ in \eqref{equ:phiqp00}, i.e., 
\begin{align}
\label{equ:phipq}
\ff^*_{q,p}(\cdot) = \E_{x\sim q}[ k(x, \cdot) \nabla_x \log p(x)  + \nabla_{x} k(x, \cdot)],
\end{align}
for which the negative gradient in \eqref{equ:pdir} equals KSD, that is, $\nabla_{\para}\KL(q_{[\T]} ~||~ p ) ~\big|_{\para = 0} =-\S(q, ~ p)$. 
%the square root of the negative kernelized Stein discrepancy, that is, $\nabla_{\para}\KL(q_{[\T]} ~||~ p ) ~\big|_{\para = 0} =  - \sqrt{\S(p, ~ q)}$. 
\end{lem}
%\red{This result has a close connection with Fisher divergence and \emph{ de Bruijn's identity}; see Appendix. }
%\blue{Note that if we take $\ff(x)= \ff^*_{q,p}(x)/||\ff^*_{q,p}||_{\H^d}$ for which $||\ff||_{\H^d} =1$, we have $\nabla_{\para}\KL(q_{[\T]} ~||~ p ) ~\big|_{\para = 0} =-\sqrt{\S(q, ~ p)}$. It is more convenient to calculate the unnormalized $\ff^*_{q,p}(x)$. }

%2. We have $$ \min_{\ff \in \H\colon ||\ff||_\H \leq 1} \bigg\{ \nabla_{\para}\KL(q_{\para\ff} ~||~ p ) ~\big|_{\para = 0}   \bigg\} =   -   \S(q_0,  ~p),$$ %\E_{x\sim q_0} [\trace(\stein \ff(x))],  and the optimal $\ff$ is obtained at %$\ff^*(\cdot) \propto $$$\ff^*(\cdot) \propto \E_{x\sim q_0}[\stein_p k(x, \cdot)]. $$
%Therefore, $\ff^*(x)$ defines the optimal smooth perturbation direction that decreases $\KL(q_\para ~||~p)$ at the fastest speed, 
The result in Lemma \eqref{lem:dd} suggests an iterative procedure that transforms 
an initial reference distribution $q_0$ to the target distribution $p$: 
%\begin{align}
%T_\ell^*(x) =  x + \epsilon \cdot \ff^*_{q_0,p}(x), 
%q_\ell(x) =  q_{t [T_\ell^*]}(x)
%\end{align}
%s
we start with applying transform $\T_0^*(x) = x + \epsilon_0 \cdot \ff^*_{q_0,p}(x)$ on $q_0$ which decreases 
 the KL divergence by an amount of $\epsilon_0 \cdot \S(q_0, ~ p)$, where $\epsilon_0$ is a small step size; % of $\epsilon$; 
 this would give a new distribution $q_1(x) = q_{0[\T_0]}(x)$, on which a further transform $\T_1^*(x) = x + \epsilon_1 \cdot \ff^*_{q_1,p}(x)$ can further decrease the KL divergence by $\epsilon_1 \cdot \S(q_1, ~ p)$. 
 Repeating this process one constructs a path of distributions $\{q_\ell\}_{\ell=1}^n$ between $q_0$ and $p$ via
\begin{align}\label{equ:iterp}
q_{\ell+1}  =  q_{\ell [\T_\ell^*]}, 
&&\text{where}&&
\T_\ell^*(x) =  x + \epsilon_\ell \cdot \ff^*_{q_\ell,p}(x). 
\end{align}
This would eventually converge to the target $p$ with sufficiently small step-size $\{\epsilon_\ell\}$, 
under which $\ff^*_{p,q_{\infty}}(x) \equiv 0$ and  $\T^*_{\infty}$ reduces to the identity map. Recall that $q_{\infty}=p$ if and only if $\ff^*_{p,q_\infty}(x)  \equiv 0$.
%\blue{Note that $q_{\infty}=p$ if and only if $\ff^*_{p,q_\infty}(x)  \equiv 0$, because $\S(q, ~ p) = || \ff^*_{q,p}||_{\H^d}^2$ as shown in \citet{liu2016kernelized}.} 
%Note that as $q_{\infty}=p$, we have $\ff^*_{p,q_\infty}(x) =0$ as a consequence of Stein's identity \eqref{equ:steq1} when applied on $k(\cdot, x')$ for fixed $x'$. 
%one should choose $\ff(x)$ to maximize the negative gradient $\E_{q}[\stein_p\ff(x)]$, and by applying $\T(x) = x + \epsilon \ff^*(x)$, we decrease the KL divergence by amount of $\epsilon \cdot \S(q, ~ p);$  
%this can be applied recursively until we arrive $p$, in which case we have $\ff^*(x) = 0$, that is, $\E_{x\sim q_0}[\stein_p k(x, \cdot)] = 0 $ for $\forall x$. 
 %according to Theorem 3.7 of \citet{liu2016kernelized}, the $\ff(x)$ that maximizes $\E_{q}[\stein_p\ff(x)]$ in the unit ball of $\H^d$ is $$\ff^*(\cdot) \propto \E_{x\sim q}[\stein_p k(x, \cdot)]. $$
%\red{[]} This suggests that we can take 
%This suggests that we can recursively apply 
%Following this gradient descent direction recursively allows us to transform the probability from $q_0$ to $p(x)$: 
%Let $x$Starting from $q_0$, we apply transform $T^*_\epsilon(x) = x + \epsilon \ff^*(x) $.
%Note that as the step size approaches to zero, we effectively follow a differential equation  ... 
%$
%\myp{q(x)}{t} = 
%$
% update 
%$T^*_\epsilon(x) = x + \epsilon \ff^*(x)$, where $\epsilon$ is a small number, to transport $q(x)$ to $p(x)$ with the fastest speed. 

\paragraph{Functional Gradient}
To gain further intuition on this process, we now reinterpret \eqref{equ:phipq} as a functional gradient in RKHS. 
For any functional $F[\vf]$ of $\vf\in \H^d$, its (functional) gradient $\nabla_\vf F[\vf]$ is a function in $\H^d$ such that $F[\vf + \epsilon \vg(x)] = F[\vf] + \epsilon ~\la \nabla_\vf F[\vf], ~ \vg \ra_{\H^d} ~+~ \Od(\epsilon^2)$
for any $\vg\in \H^d$ and $\epsilon \in \R$. 
\begin{thm}\label{thm:fungrad}
Let $\T(x) = x + \vf(x)$, where $\vf \in \H^d$, and $q_{[\T]}$ the density of $z=\T(x)$ when $x\sim q$, %we have
$$
\nabla_{\vf} \KL(q_{[\T]} ~||~ p) ~ \big | _{\vf=0} = %- \E_{x\sim q}[ \nabla_x \log p(x) k(x, \cdot) + \nabla_{x} k(x, \cdot)] \equiv 
- \ff_{q,p}^*(x), 
$$
whose squared RKHS norm is  $ || \ff_{q,p}^* ||_{\H^d}^2 = \S(q,p)$.
\end{thm}
This suggests that $\T^*(x) = x +  \epsilon \cdot \ff^*_{q,p}(x)$ is equivalent to a step of functional gradient descent in RKHS. 
However, what is critical in the iterative procedure \eqref{equ:iterp} is that we also iteratively apply the variable transform so that every time we would only need to evaluate the functional gradient descent 
 at zero perturbation $\vv f = 0$ on the identity map $\T(x)=x$. 
This brings a critical advantage since the gradient at $\vv f \neq 0$ is more complex and would require to calculate the inverse Jacobian matrix $[\nabla_{x} \T(x)]^{-1}$ that casts computational or implementation hurdles. 
%is distinguish from a regular functional gradient descent algorithm in RKHS, since 

\begin{algorithm}[tb] % 
\caption{Bayesian Inference via Variational Gradient Descent}  
\label{alg:alg1}
\begin{algorithmic}
\STATE {\bf Input:} A target distribution with density function $p(x)$ and a set of initial particles $\{x_i^0\}_{i=1}^n$. 
\STATE {\bf Output:} A set of particles $\{x_i\}_{i=1}^n$ that approximates the target distribution.  
\FOR{iteration $\ell$}
\vspace{-1.5\baselineskip}
\STATE %Select a mini-batch $\mathcal M \subset \{1,\ldots,n\}$ and $\mathcal N \subset \{1, \ldots, N\}$, update $x_i  \gets x_i  + \Delta x_i$, where
%\vspace{-1\baselineskip}
\begin{align} \label{equ:updatex}
%x_i  \gets   x_i + \epsilon_\ell \hatff{}^*(x_i) 
\!\!\!\!\!\!\! x_i^{\ell+1}  \gets   x_i^\ell + \epsilon_\ell \hatff{}^*(x_i^\ell) 
&&
\text{where}
&&
\hatff{}^*(x) = \frac{1}{n}\sum_{j=1}^n\big[  k(x_j^\ell, x)  \nabla_{x_j^\ell} \log p(x_j^\ell) + \nabla_{x_j^\ell} k(x_j^\ell, x)\big],
%& \Delta x_i   =  \epsilon_\ell \cdot   \frac{1}{|\mathcal M|}  \sum_{j \in \mathcal M} \bigg\{ \hat \score_p(x) k(x_j, x_i) + \nabla_x k(x_j, x_i) \bigg\},  \\
%& \hat \score_p(x)  =  \nabla_x \log p_0(x) + 
% \frac{N}{|\mathcal N|}\sum_{k\in \mathcal N} \nabla_x  \log p(D_k \mid x),
\end{align}
\vspace{-1\baselineskip}
\STATE where $\epsilon_\ell$ is the step size at the $\ell$-th iteration. %(e.g., as decided by AdaDelta \citep{zeiler2012adadelta}). 
\ENDFOR
\vspace{-1\baselineskip}
\STATE 
%\emph{Notes.}
%5\change{a,b,r}
%1. It is default to use the RBF kernel $k(x,x') =\exp(-\frac{1}{h}||x-x'||^2)$, with $h=\mathrm{med}^2/\log n$ where $\mathrm{med}$ is the median pairwise distance of the current points $\{x_i\}$ (which changes adaptively across the iterations). \\
%2. We recommend gradient scheme $\epsilon = a/(b+t)^{r}$ with $\blue{r=.55}$ and $a$ and $b$ ......%as suggested by \citet{bottou2012stochastic},
% or AdaDelta \citep{zeiler2012adadelta}, or AdaGrad \citep{duchi2011adaptive} to minimize the need for parameter tuning.  \\%% further automatization.\\
%4. When $N$ is very large (a.k.a. big data), use the mini-batch approximation shown in \eqref{equ:minibatch}. 
\end{algorithmic}
\end{algorithm}

\subsection{Stein Variational Gradient Descent}
To implement the iterative procedure \eqref{equ:iterp} in practice, %we start with drawing a sample $\{x_i^{(0)}\}_{i=1}^n$,  
one would need to approximate the expectation for calculating $\ff_{q,p}^*(x)$ in \eqref{equ:phipq}. 
%This is done in Algorithm~\ref{alg:alg1}, %which iteratively constructs particles $\{x_i \}$
%where
To do this,  we can first draw a set of particles $\{x_i^0\}_{i=1}^n$ from the initial distribution $q_0$, and then iteratively update the particles 
with an empirical version of the transform in \eqref{equ:iterp} in which 
the expectation under $q_\ell$ in $\ff_{q_\ell,p}^*$ is approximated by the empirical mean of particles $\{x_i^{\ell}\}_{i=1}^n$ at the $\ell$-th iteration. 
This procedure is summarized in  Algorithm~\ref{alg:alg1}, 
which allows us to (deterministically) transport a set of points to match our target distribution $p(x)$, effectively providing a sampling method for $p(x)$. 
We can see that this procedure does not depend on the initial distribution $q_0$ at all, meaning that we can apply this procedure starting with  a set of arbitrary points $\{x_i\}_{i=1}^n$, possibly generated by a complex (randomly or deterministic) black-box procedure. 
%which forms a \emph{deterministic} procedure that

We can expect that $\{x_i^\ell\}_{i=1}^n$ forms increasingly better approximation for $q_\ell$ as $n$ increases. 
To see this, denote by $\Phi$ the nonlinear map that takes the measure of $q_\ell$ and outputs that of $q_{\ell + 1}$ in \eqref{equ:iterp}, that is, $q_{\ell + 1} = \Phi_\ell(q_\ell)$,
where $q_\ell$ enters the map through both $q_{\ell[\T_\ell^*]}$ and $\ff^*_{q_\ell, p}$. 
Then, the updates in Algorithm~\ref{alg:alg1} can be seen as applying the same map $\Phi$ 
on the empirical measure $\hat q_{\ell}$ of particles $\{x_i^\ell\}$ to get the empirical measure $\hat q_{\ell + 1}$ of particles $\{x_i^{\ell+1}\}$ at the next iteration, that is, $\hat q_{\ell + 1} = \Phi_\ell(\hat q_\ell)$. 
%except with the inputs replaced by $\hat q_{\ell}$, the empirical measure of the particles $\{x_i^\ell\}$, that is,  $\hat q_{\ell + 1} = \Phi_\ell(\hat q_\ell)$. 
%corresponds to $\hat q_{\ell + 1} = \Phi_\ell(\hat q_\ell)$, where $\hat q_{\ell} = \sum_i \delta(x - x_i)/n$, 
%Therefore
Since $\hat q_0$ converges to $q_0$ as $n$ increases, $\hat q_\ell$ should also converge to $q_\ell$ when the map $\Phi$ is ``continuous'' in a proper sense. %regular. 
Rigorous theoretical results on such convergence have been established
in the mean field theory of interacting particle systems \citep[e.g.,][]{del2013mean}, which in general guarantee that %the empirical measures of the particles $\{x_i^\ell\}$ weakly converge to the measure of $q_\ell$ with an fluctuation of order $1/\sqrt{n}$, in the sense that 
$\sum_{i=1}^n h(x_i^\ell)/n - \E_{q_\ell} [h(x)] = \bigO{1/\sqrt{n}}$ for bounded testing functions $h$. 
%%, under certain regularity conditions on the mapping $\Phi_\ell$, the empirical measures of the particles 
%$\{x_i^\ell\}$ weakly $\{x_i^\ell\}$ weakly converge to the measure of $q_\ell$ with an fluctuation of order $1/\sqrt{n}$, in the sense that $\sum_{i=1}^n h(x_i^\ell)/n - \E_p [h(x)] = \bigO{1/\sqrt{n}}$ for any continuous bounded function $h$ . 
In addition, the distribution of each particle $x_{i_0}^\ell$, for any fixed $i_0$, also tends to $q_\ell$, and is independent with any other finite subset of particles as $n\to \infty$, a phenomenon  called \emph{propagation of chaos} \citep{kac1959probability}.
%Interesting connections can also be drawn with Vlasov equation in kinetic theory. 
We leave concrete theoretical analysis for future work.

Algorithm~\ref{alg:alg1}  mimics a gradient dynamics at the particle level, %that drives the particles to match the target distribution, 
%the two terms of in the functional gradient which the two terms in $\hat \ff^*(x)$ in \eqref{equ:updatex} yield intuitive interpretation: 
where the two terms in $\hatff{}^*(x)$ in \eqref{equ:updatex} play different roles:  
the first term drives the particles towards the high probability areas of $p(x)$ by following a \emph{smoothed} gradient direction, which is the weighted sum of the gradients of all the points weighted by the kernel function. 
The second term acts as a \emph{repulsive force} that prevents all the points to collapse together into local modes of $p(x)$; 
to see this, consider the RBF kernel $k(x,x') = \exp(-\frac{1}{h}||x - x'||^2)$, the second term reduces to $\sum_{j} \frac{2}{h}(x- x_j)k(x_j,x)$, which drives $x$ away from its neighboring points $x_j$ that have large $k(x_j, x)$.  
\blue{
%Note that i
If we let bandwidth $h \to 0$, the repulsive term vanishes, and update \eqref{equ:updatex} reduces to a set of independent chains of
typical gradient ascent for maximizing $\log p(x)$ (i.e., MAP) and all the particles would collapse into the local modes. % of $p(x)$. 
}
%\todo{Algorithm~\ref{alg:alg1} approximates the (nonlinear) density evolution in \eqref{equ:iterp} with a interacting particle system.} 

%\blue{
Another interesting case is when we use only a single particle ($n = 1$), 
in which case Algorithm~\ref{alg:alg1} reduces to a single chain of typical gradient ascent for MAP for any kernel that satisfies $\nabla_x k(x, x) =0$ (for which RBF holds). 
%We should remark that the practical efficiency of our method does not critical rely on the large $n$ limit, 
%this is because Algorithm~\ref{alg:alg1} reduces to the typical gradient ascent for MAP with only a single particle ($n=1$), 
This suggests that our algorithm can generalize well for supervised learning tasks even with a very small number $n$ of particles, 
since gradient ascent for MAP ($n=1$) has been shown to be very successful in practice. 
%which has been widely used in practice and achieve good prediction performance. 
This property distinguishes our particle method with the typical Monte Carlo methods that requires to average over many points. % to obtain good results.  
The key difference here is that we use a deterministic repulsive force, other than Monte Carlo randomness, to get diverse points for distributional approximation. 

\todo{
Our algorithm works for generic unnormalized distribution once $\nabla_x \log p(x)$ is provided. 
It can also applied on distributions that are weakly differentiable. %, in the sense that Stein's identity \eqref{equ:} , once the weak derivative is provided. 
For example, the weak derivative of RELU is $\max(0, x)$ is $\ind[x>=0]$. 
{see: \url{https://www.dropbox.com/s/1ef8mirsl67n7gm/livrable_2_3.pdf?dl=0}}
}

%This process continues util $\{x_i\}_{i=1}^n$ closely match with $p(x)$ such that $\frac{1}{n}\sum_{j=1}^n \stein_p k(x_j, x) \approx 0$. 

%\textbf{Conve}
%The idea of approximating density evolutions with particle schemes have been widely ad
%On the other hand, 
%we rewrite the transform in \eqref{equ:iterp} as $q_{\ell + 1} = \Phi_\ell(q_\ell)$, where $\Phi_\ell$ 
%which forms an increasingly better approximation for $q_\ell$ as $n$ increases. 
%In this way, $\{x_i^\ell\}_{i=1}^n$ forms an increasingly better approximation for $q_\ell$ as $n$ increases, 
%This is formally guaranteed by the mean field theory of interacting particle systems \citep[e.g.,][]{del2013mean}, which in general states that the empirical measures of the particles $\{x_i^\ell\}$ weakly converge to the measure of $q_\ell$ with an fluctuation of order $1/\sqrt{n}$, in the sense that $\sum_{i=1}^n h(x_i^\ell)/n - \E_p [h(x)] = \bigO{1/\sqrt{n}}$ for any continuous bounded function $h$. 
%In addition, the distribution of each particle $x_{i_0}^\ell$, for any fixed $i_0$, also tends to $q_\ell$, and is independent with any other finite subset of particles as $n\to \infty$, a phenomenon  called \emph{propagation of chaos} \citep{kac1959probability}.

\paragraph{Complexity and Efficient Implementation}
The major computation bottleneck in \eqref{equ:updatex} lies on calculating the gradient $\nabla_{x}\log p(x)$ for all the points $\{x_i\}_{i=1}^n$; 
%which is linear on $n$ and also depends on the complexity of $p(x)$. 
this is especially the case in big data settings when $p(x) \propto p_0(x) \prod{}^N_{k=1} p( D_k | x)$ with a very large $N$.  
We can conveniently address this problem by %using the stochastic gradient trick, approximates $\nabla_x \log p(x)$ with 
approximating $\nabla_x \log p(x)$ with subsampled mini-batches $\Omega \subset \{1,\ldots, N\}$ of the data%, and approximates %$\nabla_x \log p(x)$ with 
\begin{align}\label{equ:minibatch}
% {\nabla_x \log p(x)} \approx \log p_0(x) + \frac{N}{|\Omega|}\sum_{k\in \Omega} \log p(D_k \mid x). 
\log p(x) \approx \log p_0(x) + \frac{N}{|\Omega|}\sum_{k\in \Omega} \log p(D_k \mid x). 
\end{align}
%This allows us to efficiently apply our algorithm to big data problems. 
Additional speedup can be obtained by parallelizing the gradient evaluation of the $n$ particles.
\fullversion{We can further approximate $p(D_i \mid x)$ with $\hat p_i$, and sum over $\log p(D_i \mid x)  - \log \hat p_i $.}
% where  $p_k(x) \propto p_0(x)^{1/N} p(D_k | x )$, and hence $\nabla_x \log p(x) = \sum_{k=1}^N \log p_k(x)$. When $N$ is very large (that is, big data), we can further approximate the RHS of \ref{equ:updatex} by susampling the $\sum_{i=1}^N$. 

The update \eqref{equ:updatex} also requires to compute the kernel matrix $\{k(x_i,x_j)\}$ which costs $\bigO{n^2}$; in practice, this cost can be relatively small compared with the cost of gradient evaluation, 
since it can be sufficient to use a relatively small $n$ (e.g., several hundreds) in practice. % which eliminates the need for further approximation. 
If there is a need for very large $n$, one can approximate the summation $\sum_{i=1}^n$ in \eqref{equ:updatex} by subsampling the particles, 
or using a random feature expansion of the kernel $k(x,x')$ \citep{rahimi2007random}. 

\section{Related Works}
\label{sec:related}
Our work is mostly related to \citet{jimenez2015variational}, 
which also considers variational inference over the set of transformed random variables, 
but focuses on transforms of parametric form $T(x) = f_\ell( \cdots (f_1(f_0(x))))$ where $f_i(\cdot)$ is a predefined simple parametric transform and
$\ell$ a predefined length; %, and optimizes the parameters to minimize the variational KL divergence; 
this essentially creates a feedforward neural network with $\ell$ layers, whose invertibility requires further conditions on the parameters and need to be established case by case. 
The similar idea is also discussed in \citet{marzouk2016introduction}, which also considers transforms parameterized in special ways 
to ensure the invertible and the computational tractability of the Jacobian matrix. 
Recently, 
\citet{tran2015variational} constructed a  variational family that achieves universal approximation based on Gaussian process (equivalent to a single-layer, infinitely-wide neural network), which does not have a Jacobian matrix but needs to calculate the inverse of the kernel matrix of the Gaussian process.  
%\citet{tran2015variational} leverages Gaussian process (GP), equivalent to single-layer, infinitely-wide neural networks, as a variational family to achieve universal approximation, which does have a Jacobian matrix but need to calculate the inverse of the kernel matrix in GP. 
%leverages Gaussian process (GP), equivalent to single-layer, infinitely-wide neural networks, as a variational family to achieve universal approximation, which does have a Jacobian matrix but need to calculate the inverse of the kernel matrix in GP. 
%\citet{tran2015variational} leverages Gaussian process (GP), equivalent to single-layer, infinitely-wide neural networks, as a variational family to achieve universal approximation, which does have a Jacobian matrix but need to calculate the inverse of the kernel matrix in GP. 
%but requires to calculate matrix inverse
%corresponding to variable transformation using a  an one-lay, but (infinite) wide neural network, 
%More from the perspective of generating samples, 
%The proposed algorithm is related sophistic,  complicated and unaccessible to non-experts, without the goal of obtaining state-of-the-art results, builded on the basis of \citep[e.g.,][]{kingma2013auto, mnih2014neural}. 
Our algorithm has a simpler form, and does not require to calculate any matrix determinant or inversion. 
%The algorithms in \citet{jimenez2015variational} is complicated and unaccessible to non-experts, without the goal of obtaining state-of-the-art results, builded on the basis of \citep[e.g.,][]{kingma2013auto, mnih2014neural}. 
%deep learning problems, 
%while our algorithm features its simplicity, eliminating the needs for hand tuning. 
%much simpler and 
%is meant to constr 
% $T_\ell(x) = x + u h(w^\top x + b),$
%
Several other works also leverage variable transforms in variational inference, but with more limited forms; examples include affine transforms \citep{titsias2014doubly, challis2012affine}, and recently the copula models that correspond to element-wise transforms over the individual variables 
\citep{han2015variational, tran2015copula}.
%More broadly, the idea of variable transform has been the key component of many other inference and learning algorithm,  such as \citep{meng2002warp, hauberg2015dreaming}, to name only a few. 
%\red{There are many other inspiring works that uses variable transforms outside of variational inference,  such as, \citep{meng2002warp, hauberg2015dreaming}, to name only a few. }
%for example,  \citet{titsias2014doubly, challis2012affine} used affine transforms of simple distributions for approximation,  and \citet{han2015variational, tran2015copula} 
%\red{Interesting connections can also be drawn between our method and Stein's identity and the reparameterization trick \citep[e.g.,][]{kingma2013auto, mnih2014neural}, which will be explored in future works. }
%Further connection will be explored in future works. }
%which can also be derived using integration by parts.  Further connection will be explored in future works.} 
%\red{Our method and Stein's identity is closely related to the reparameterization trick \citep[e.g.,][]{kingma2013auto, mnih2014neural} which can also be derived using integration by parts.  Further connection will be explored in future works.} 
%\todo{Recently, neural variational inference \citep[e.g.,][]{kingma2013auto, mnih2014neural} assumes $\mathcal Q$ is parameterized by neural networks, which then introduces too much complexity and may not worthwhile for simple models. }

Our algorithm maintains and updates a set of particles, 
and is of similar style with the Gaussian mixture variation inference methods whose mean parameters can be treated as a set of particles. 
 \citep{lawrence1998approximating, jaakkola1999improving, lawrence2001variational, kulkarni2014variational, gershman2012nonparametric}.
%and is similar to nonparametric variational inference \citet{gershman2012nonparametric}
%which uses a Gaussian mixture as the variational distribution; 
%see also \citet{lawrence1998approximating, jaakkola1999improving, lawrence2001variational}.  
Optimizing such mixture KL objectives often requires certain approximation, 
and this was done most recently in \citet{gershman2012nonparametric} by approximating the entropy using Jensen's inequality and the expectation term using Taylor approximation. 
%\citet{kulkarni2014variational} instead uses a weighted combination of particles (corresponds to )
%mixture of delta functions to 
There is also a large set of particle-based Monte Carlo methods, including variants of  
sequential Monte Carlo \citep[e.g.,][]{robert2013monte,smith2013sequential}, 
as well as a recent particle mirror descent for optimizing the variational objective function \citep{dai2016provable};
compared with these methods, our method does not have the weight degeneration problem, and is much more ``particle-efficient'' in that we reduce to MAP with only one single particle. %s\red{particle MCMC?}
%compared with our method which updates the particles by gradient descent, these methods update the particles by \emph{reweighing} and \emph{resampling}, and may suffer from the weight degeneration problem. 

%Another particle-based variational inference is recently developed by \citet{dai2016provable} 
%which uses importance weighted kernel density estimator to represent $q(x)$. 
%\gray{Also related is a mirror descent algorithm that uses importance weighted kernel density estimation on the variational distribution \citep{dai2016provable}}. 
%Finally, there is a large set of particle-based Monte Carlo methods, such as sequential particle filtering and population Monte Carlo; see e.g., \citet{robert2013monte,smith2013sequential} for overviews.  
%on the Monte Carlo side, 
%Traditionally $\mathcal Q$ is taken to be the set of mean field models, or Gaussian models, which restricts its flexibility and introduces large approximation errors. 
%Highly related to our work is the Gaussian mixture variational inference, which however experiments difficulty in calculating the KL objective \citep{lawrence1998approximating, jaakkola1999improving, lawrence2001variational}; this was recently done by approximating the entropy using Jensen's inequality and the expectation term using Taylor approximation \citep{gershman2012nonparametric}. 
%We are also aware of methods use affine transforms of simple distributions for approximation \citep{titsias2014doubly, challis2012affine}. 
%
%Another recent mirror descent algorithm .... \citep{dai2016provable}.

\section{Experiments}
\label{sec:experiments}
%
%We show that our algorithms tends to outperform most 
We test our algorithm on both toy and real world examples, on which we find our method tends to 
outperform a variety of baseline methods. % widely used in the literature. 
%show that 
%find that  our method is comparable, or better than all the baseline algorithms that we compared with. 
%We test our algorithm on a Gaussian toy example, real-world models including Bayesian logistic regression and Bayesian neural network,
%on which we find find that  our method is comparable, or better than all the baseline algorithms that we compared with. 
Our code is available at {\url{https://github.com/DartML/Stein-Variational-Gradient-Descent}}.
%We compared with a large set of baseline algorithms and we%demonstrate the efficiency of our method using empirical results.
%We perform empirical studies and  find that  our method is comparable, or better than most baseline algorithms. In particular, we apply our method on Bayesian neural network and find it outperforms 
%one of the state-of-the-art 
%a baseline method provided by \citet{hernandez2015probabilistic}. 

For all our experiments, we use RBF kernel $k(x,x') =\exp(-\frac{1}{h}||x-x'||_2^2)$, and take the bandwidth to be $h = \mathrm{med}^2/\log n$, where $\mathrm{med}$ is the median of the pairwise distance between the current points $\{x_i\}_{i=1}^n$; this is based on the intuition that we would have $\sum_j k(x_i,x_j) \approx n \exp(-\frac{1}{h} \mathrm{med}^2) = 1$, so that for each $x_i$ the contribution from its own gradient and the influence from the other points balance with each other. Note that in this way, the bandwidth $h$ actually changes adaptively across the iterations. 
% which eliminates the need for hand tune $h$ and also allow it to changes adaptively across the iterations. 
We use AdaGrad for step size and initialize the particles using the prior distribution unless otherwise specified. 
\todo{For the gradient step size, we use AdaGrad we find that $\epsilon_\ell = a/(b+t)^{r}$ with $\red{r=.55}$ works well, and we set $a$ and $b$ by.....; more automatic options include AdaDelta \citep{zeiler2012adadelta}, or AdaGrad \citep{duchi2011adaptive} also works well????}

\begin{figure}[tbp]
   \centering
   \begin{tabular}{cccccc}
   \includegraphics[height=.15\textwidth, trim={0cm 0 0 0}, clip]{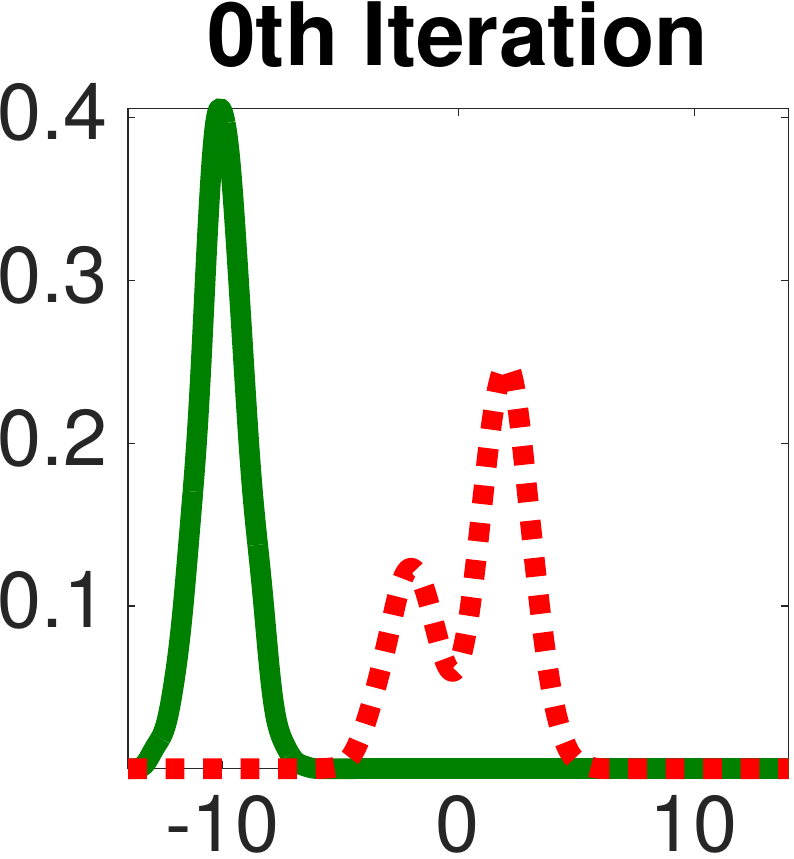} &
   \includegraphics[height=.15\textwidth, trim={1.99cm 0 0 0}, clip]{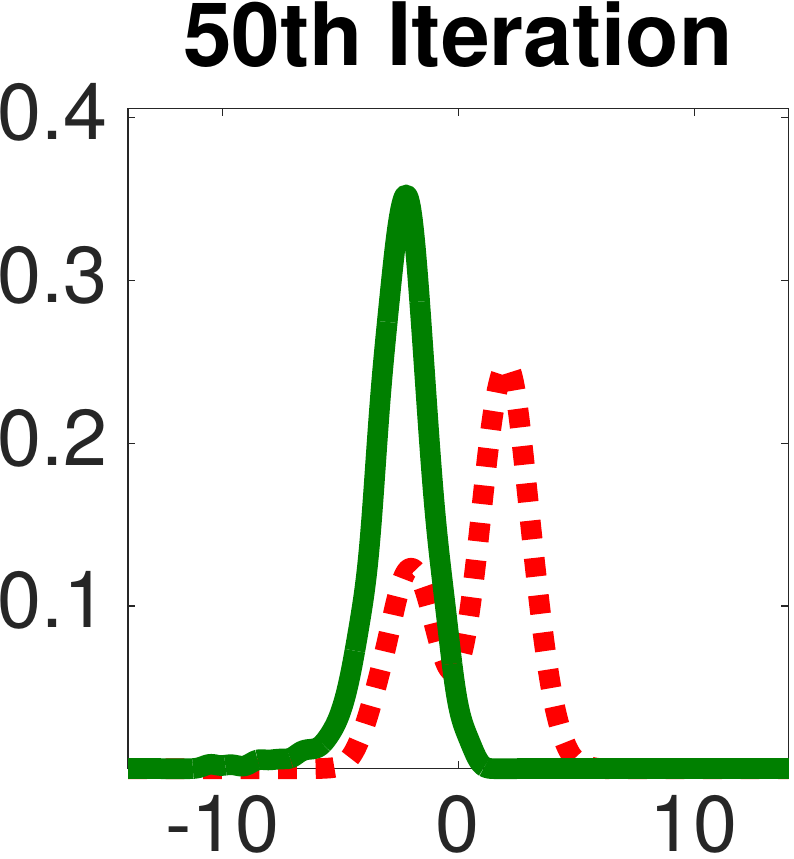} &
   \includegraphics[height=.15\textwidth, trim={1.99cm 0 0 0}, clip]{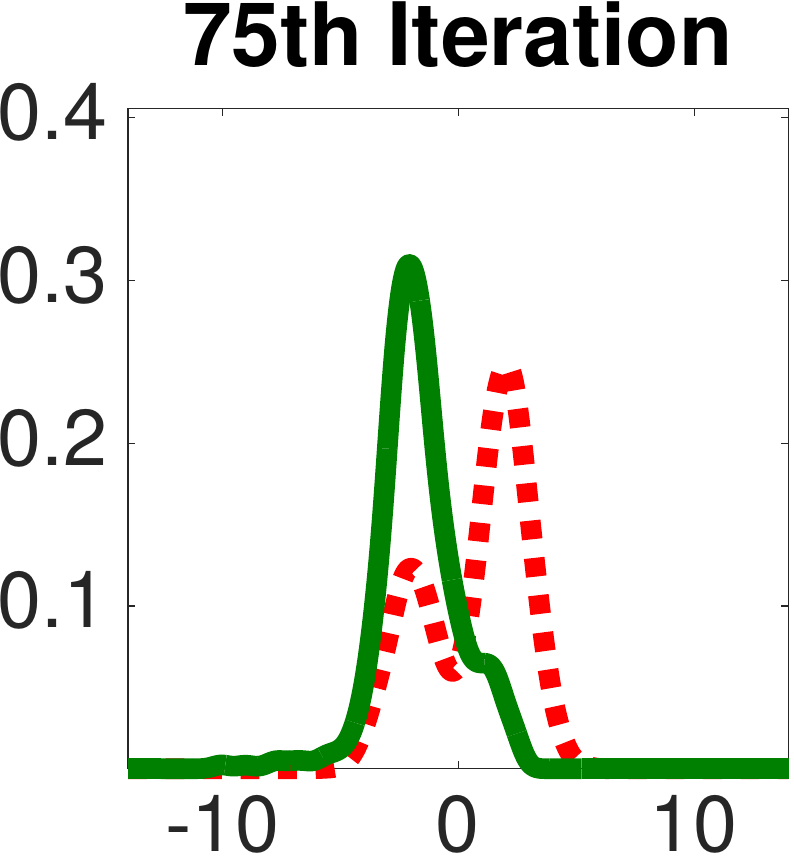} &
   \includegraphics[height=.15\textwidth, trim={1.99cm 0 0 0}, clip]{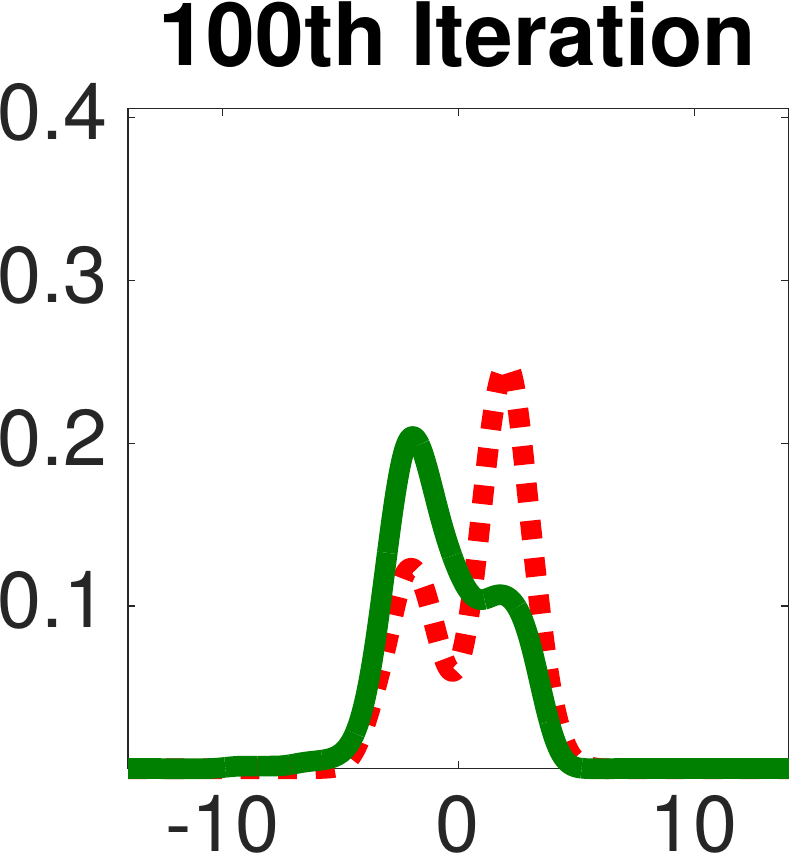} &
   \includegraphics[height=.15\textwidth, trim={1.99cm 0 0 0}, clip]{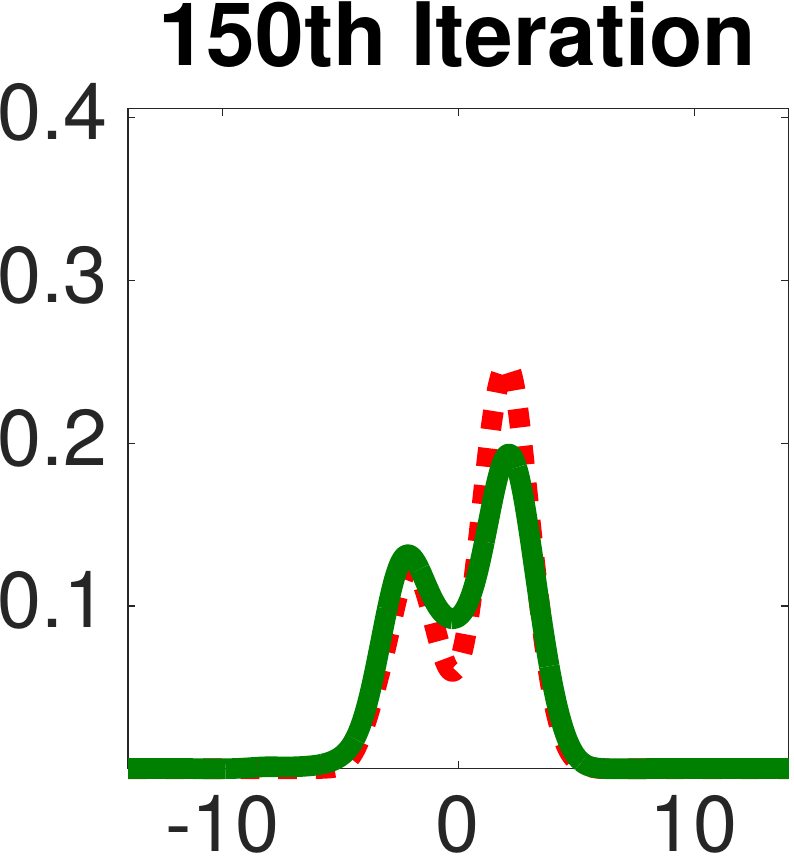}  &   
   \includegraphics[height=.15\textwidth, trim={1.99cm 0 0 0}, clip]{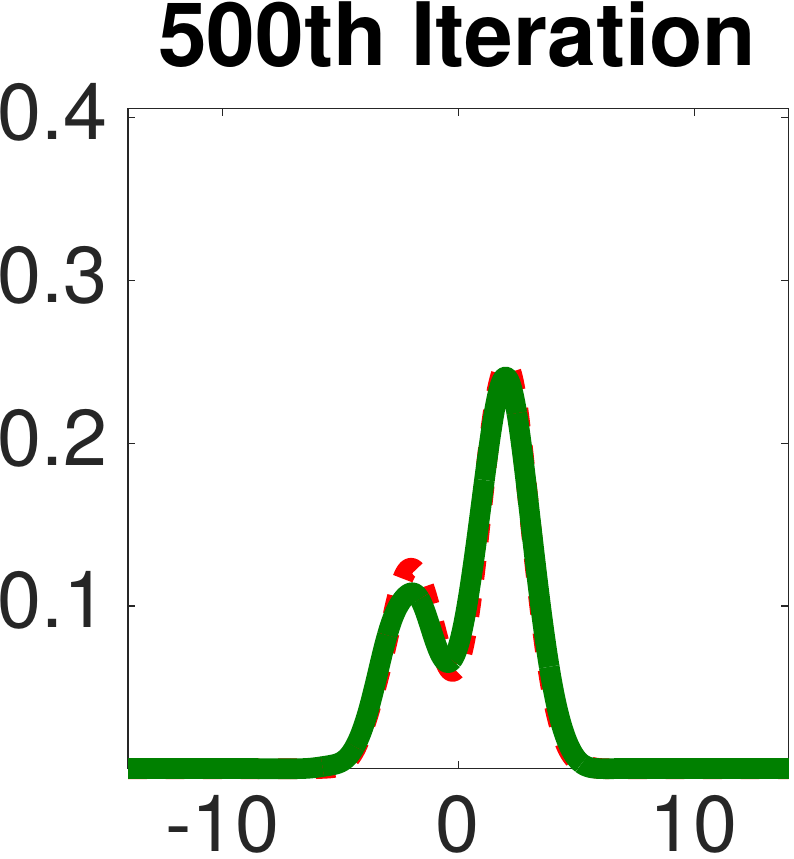}  
   \end{tabular}
   \caption{Toy example with 1D Gaussian mixture. The red dashed lines are the target density function and the solid green lines are the densities of the particles at different iterations of our algorithm (estimated using kernel density estimator) . Note that the initial distribution is set to have almost zero overlap with the target distribution, and our method demonstrates the ability of escaping the local mode on the left to recover the mode on the left that is further away. We use $n=100$ particles.}
   \label{fig:1dgmm1}
\end{figure}

%\subsection{Toy Example on 1D Gaussian Mixture}
\paragraph{Toy Example on 1D Gaussian Mixture}
%We start with a toy example with a 1D Gaussian mixture. 
We set our target distribution to be $p(x)=1/3\normal(x ;~ -2,1) +2/3 \normal(x; ~ 2, 1)$, and initialize the particles using 
$q_0(x)=\normal(x; -10,1)$. This creates a challenging situation since the probability mass of $p(x)$ and $q_0(x)$ are far away each other (with almost zero overlap). 
Figure~\ref{fig:1dgmm1} shows how the distribution of the particles $(n=1)$ of our method evolve at different iterations. %when using $n=100$ particles. 
We see that despite the small overlap between $q_0(x)$ and $p(x)$, our method can push the particles towards the target distribution, 
and even recover the mode that is further away from the initial point. 
\blue{We found that other particle based algorithms, such as  \citet{dai2016provable}, tend to experience weight degeneracy on this toy example due to the ill choice of $q_0(x)$.}
%We also tested the particle mirror descent 
%note that a typical optimization algorithm with the same initializations would mostly converge on the mode on the left, and fail to find the larger mode on the right. 

\begin{figure}[tbp]
   \centering
   \begin{tabular}{lllc}
   \includegraphics[height=.14\textwidth, trim={0cm 0 0 0}, clip]{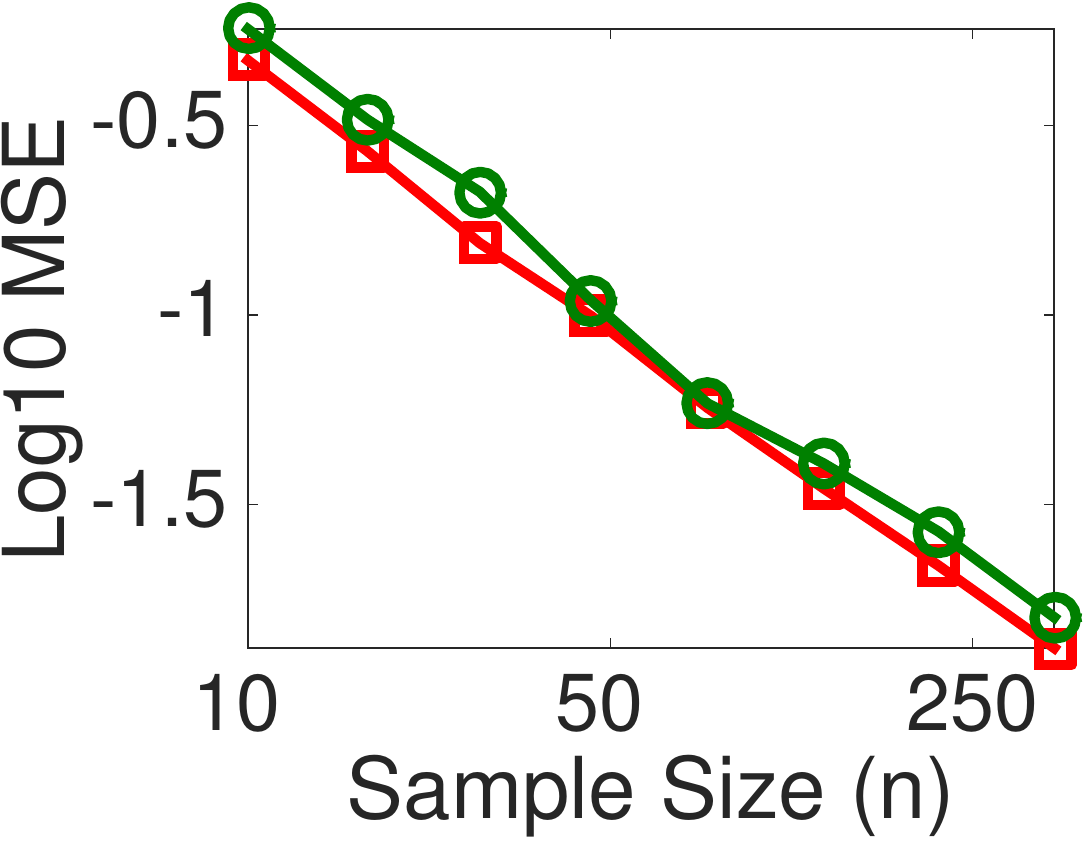} &
   \includegraphics[height=.14\textwidth, trim={1.4cm 0 0 0}, clip]{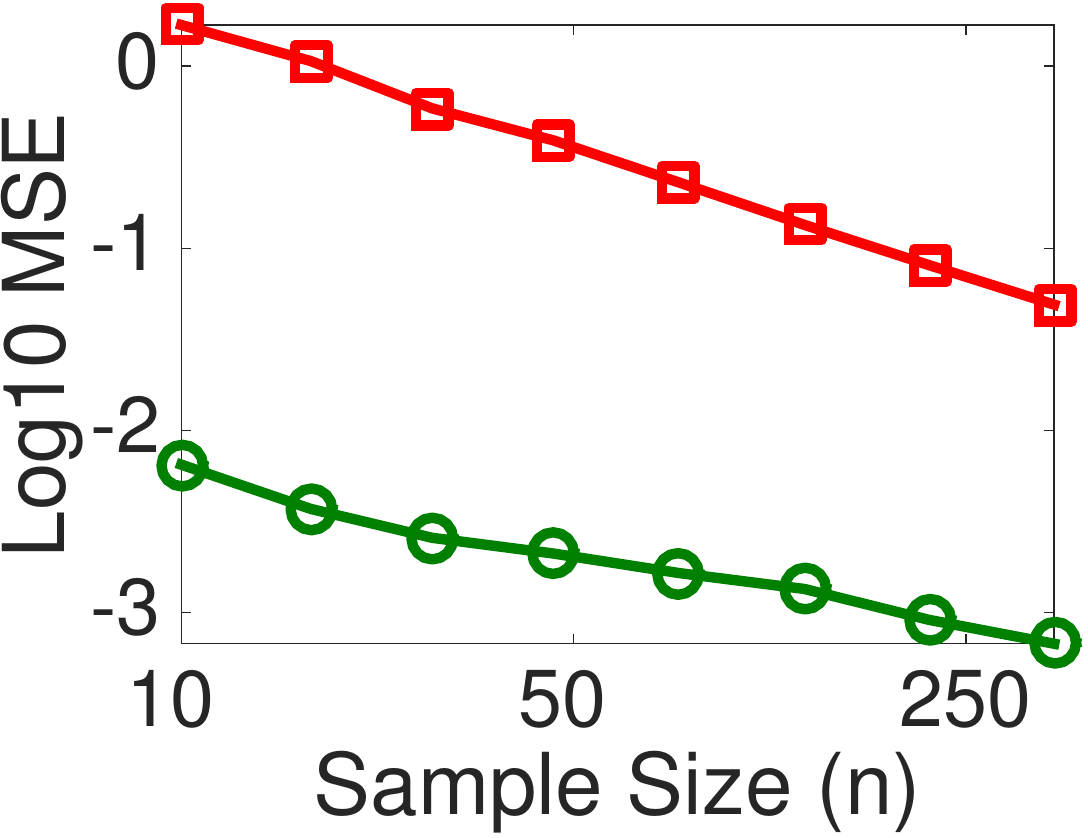} &
   \includegraphics[height=.14\textwidth, trim={1.99cm 0 0 0}, clip]{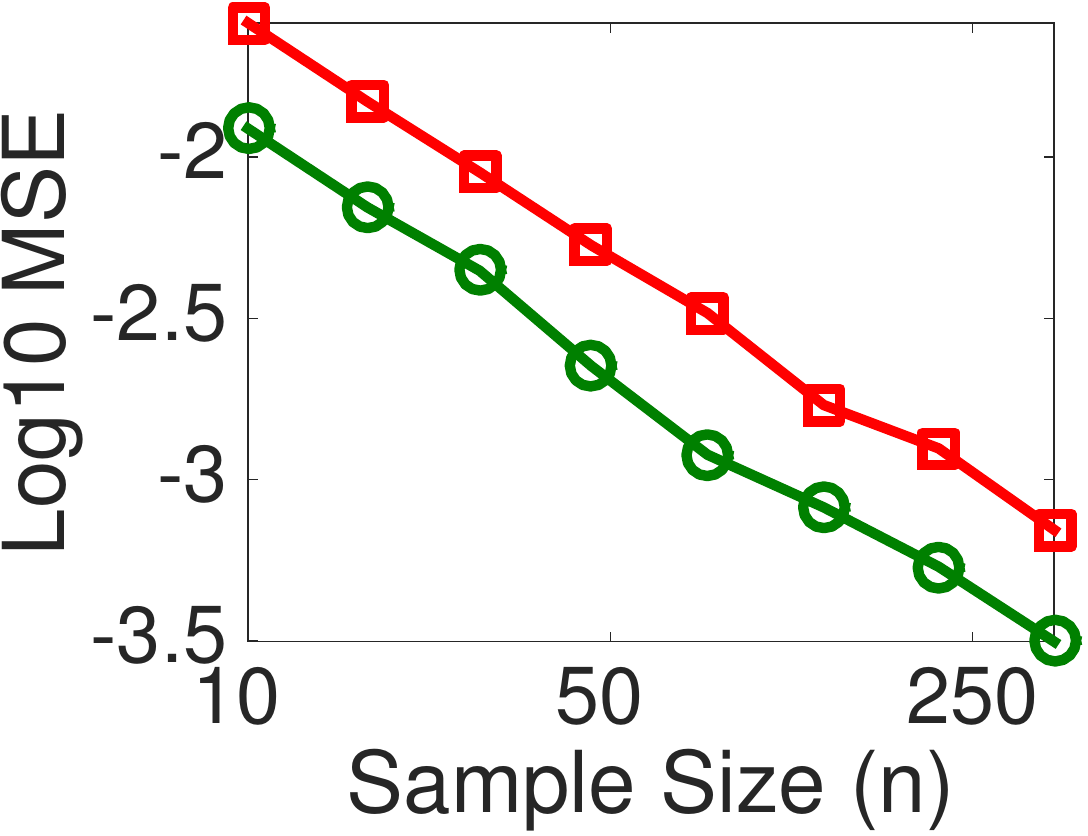} &
\hspace{-3em}\raisebox{4.0em}{   \includegraphics[height=.04\textwidth, trim={0 0 0 0}, clip]{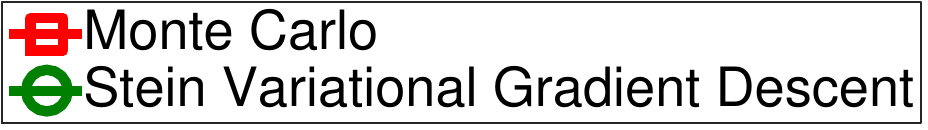}    } \\
{\small\it (a) Estimating $\E(x)$} & 
{\small\it(b) Estimating $\E(x^2)$} & 
{\small\it (c) Estimating $\E(\cos(\omega x+b))$}  &
   \end{tabular}
   \caption{We use the same setting as Figure~\ref{fig:1dgmm1}, except varying the number $n$ of particles. (a)-(c) show the mean square errors when using the obtained particles to estimate expectation $\E_p(h(x))$ for $h(x)=x$, $x^2$, and $\cos(\omega x+ b)$;  for $\cos(\omega x+ b)$, we random draw $\omega\sim \normal(0,1)$ and $b\sim \mathrm{Uniform}([0,2\pi])$ and report the average MSE over $20$ random draws of $\omega$ and $b$.}
   \label{fig:1dgmm2}
\end{figure}

Figure~\ref{fig:1dgmm2} compares our method with Monte Carlo sampling when using the obtained particles to estimate expectation $\E_p(h(x))$ with different test functions $h(\cdot)$. 
\todo{for $h(x)=x$, $x^2$, and $\cos(\omega x+ b)$ with random drawn $\omega\sim \normal(0,1)$ and $b\sim \mathrm{Uniform}([0,2\pi])$ (result averaged on 20 trials). }
%We use the same setting as Figure~\ref{fig:1dgmm1}, but vary the number $n$ of particles. (a)-(c) show the mean square error when using the obtained particles to estimate expectation $\E_p(h(x))$ for $h(x)=x$, $x^2$, and $\cos(\omega x+ b)$ with random drawn $\omega\sim \normal(0,1)$ and $b\sim \mathrm{Uniform}([0,2\pi])$ (result averaged on 20 trials). 
We see that the MSE of our method tends to perform similarly or better than the exact Monte Carlo sampling.  
\blue{This may be because our particles are more spread out than i.i.d. samples due to the repulsive force, and hence give higher estimation accuracy.}
%\blue{This may be because our particles are negatively correlated with each other due to the repulsive force in the gradient update, and hence give lower estimation variance. }
%We think that our method p
%Note that the points produced by our algorithm are not independent because of the coupled gradient updates; 
%they may instead be negatively correlated with each other, which explains its better performance in Figure~\ref{fig:1dgmm2}(b)-(c). 
%and seems to decay with the same Monte Carlo $O(n^{-1/2})$ rate. 
It remains an open question to formally establish the error rate of our method. % for estimating expectations. 

\cutspace{
\begin{figure}[htbp]
\label{fig:uncer}
   \centering
   \begin{tabular}{cccc}
   \includegraphics[height=.25\textwidth, trim={0cm 0 0 0}, clip]{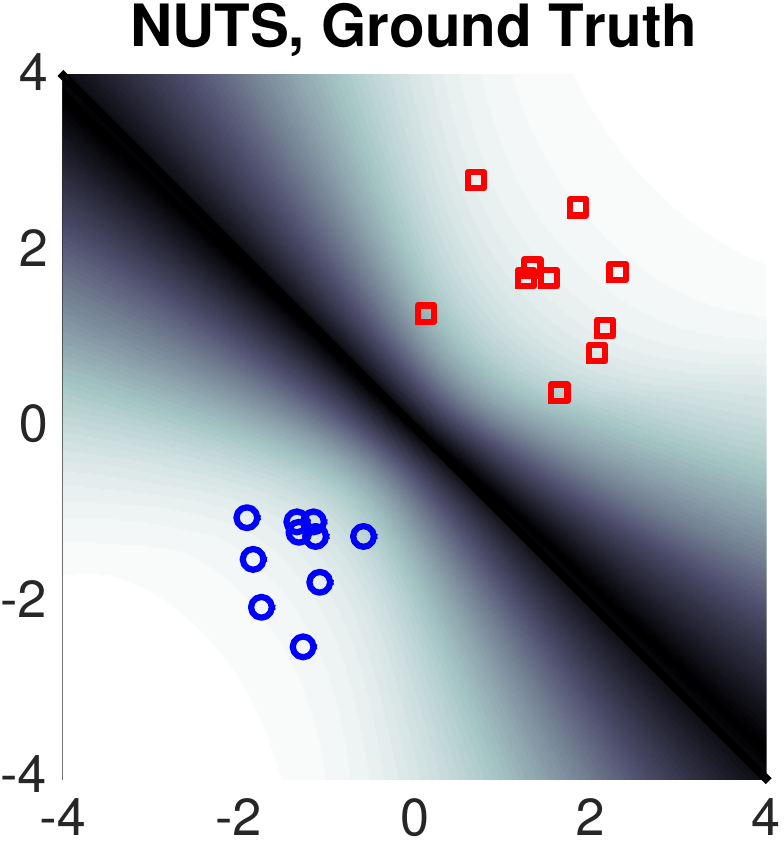} &
   \includegraphics[height=.25\textwidth, trim={0cm 0 0 0}, clip]{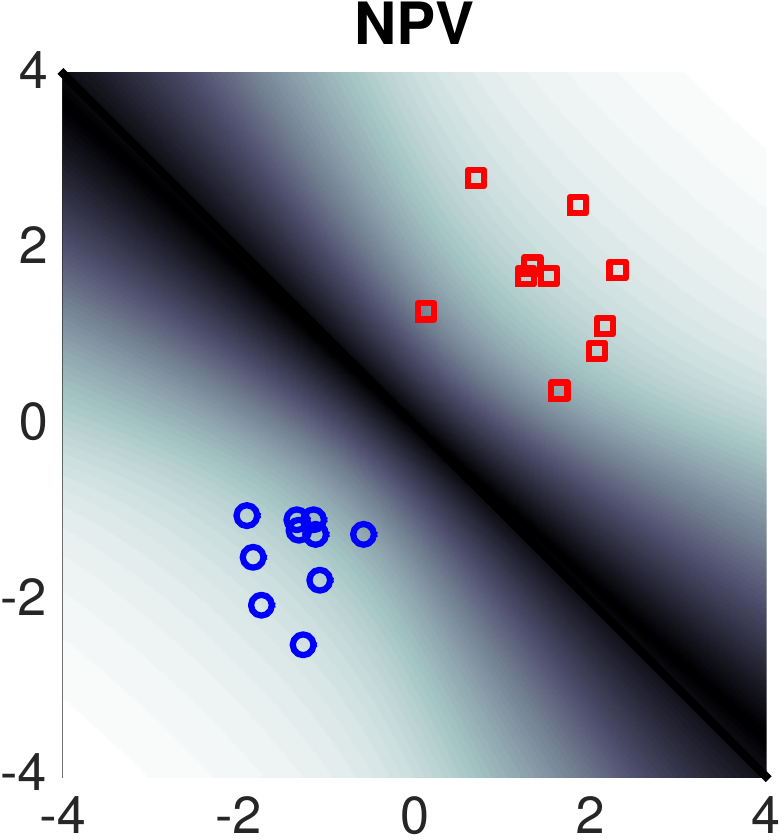} &
   \includegraphics[height=.25\textwidth, trim={0cm 0 0 0}, clip]{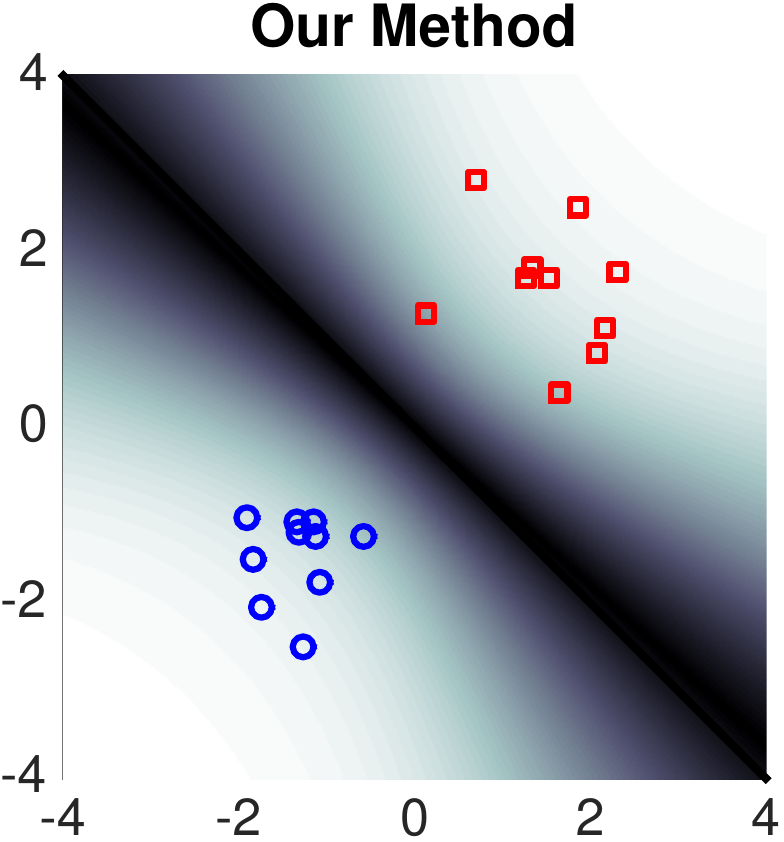} &
   \\
   \end{tabular}
   \caption{Bayesian logistic regression. The posterior prediction uncertainty as inferred by different approaches on a toy data.}
\end{figure}
}
\todo{Uncertainty: Stepsize, adagrad without momentum. master size = 5e-2}
\todo{Uncertainty: Number of particles = 100}
\todo{Uncertainty: NPV. GMM, number of components = 100}
%^
%
%
%
%
%\subsection{Bayesian Logistic Regression}
\paragraph{Bayesian Logistic Regression}
We consider Bayesian logistic regression for binary classification
using the same setting as \citet{gershman2012nonparametric}, which assigns the regression weights $w$ with a Gaussian prior $p_0(w | \alpha) = \normal(w, \alpha^{-1})$
and $p_0(\alpha) = Gamma(\alpha, 1, 0.01)$. The inference is applied on posterior $p(x | D)$ with $x = [w, \log \alpha]$. 
%The hyper-parameter is taken to be $a=1$ and $b = 0.01$. 
%Our setting is the same as that in \citet{gershman2012nonparametric}, and 
We compared our algorithm with the no-U-turn sampler (NUTS)\footnote{code: http://www.cs.princeton.edu/~mdhoffma/} \citep{homan2014no} and 
non-parametric variational inference (NPV)\footnote{code: http://gershmanlab.webfactional.com/pubs/npv.v1.zip}  \citep{gershman2012nonparametric}
 on the 8 datasets ($N>500$) used in \citet{gershman2012nonparametric}, and find they tend to give very similar results on these (relatively simple) datasets; see Appendix for more details. 
\cutspace{
We consider the Bayesian logistic regression model for binary classification, on which the regression weights $w$ is assigned with a Gaussian prior $p_0(w | \alpha) = \normal(w, \alpha^{-1})$
and $p_0(\alpha) = Gamma(\alpha, a, b)$, and apply inference on posterior $p(x \cd D)$, where $x = [w, \log \alpha]$. The hyper-parameter is taken to be $a=1$ and $b = 0.01$. 
This setting is the same as that in \citet{gershman2012nonparametric}. 
We compared our algorithm with  the no-U-turn sampler (NUTS)\footnote{code: http://www.cs.princeton.edu/~mdhoffma/} \citep{homan2014no} and 
non-parametric variational inference (NPV)\footnote{code: http://gershmanlab.webfactional.com/pubs/npv.v1.zip} on the 8 datasets as used in \citet{gershman2012nonparametric}, in which we use \red{$100$} particles,
NPV uses \red{100} mixture components, and NUTS uses \red{1000} draws with \red{$1000$} burnin period. 
We find that all these three algorithms almost always performs the same across the \red{8} datasets (See Figure ~ in Appendix), and this is consistent with Figure 2 of \citet{gershman2012nonparametric}. 
We further experimented on a toy dataset with only two features and visualize the prediction probability of the three algorithms in Figure~\ref{fig:uncer}. 
}
% compare NUTS, NPV and our method on a toy dataset with only two features. 
%For our algorithm
%
%We test our methods on Bayesian logistic regression model: 
%We applied our gradient population algorithm to a Bayesian logisitc regression model.
%
%The observed data $D = \{X; Y\}$ consist $N$ binary class labels, $y_i \in\{-1,+1\}$, with the corresponding input vector $x_i$. The target distribution is
%\begin{align*}
%p(w, \alpha \mid X, Y) &\propto p(Y \mid X, w) p(w \mid \alpha) p(\alpha) 
%\end{align*}
%where $w$ and $\alpha$ are the parameters, and $ p(y_i \mid x_i, w) = \frac{1}{ 1 + \exp(- y_i w^\top x_i)}, p(w_k \mid \alpha) = \mathcal{N}(w_k; 0, \alpha^{-1}), 
% p(\alpha)  = \mathrm{Gamma}(\alpha; a, b) $.
%The bias parameter is absorbed into $w$ by including 1 as an entry in $x_i$. 
%Here $a$ and $b$ are hyperparameters that we assume to be fixed.
%This is the same definition as in \citet{gershman2012nonparametric}.

\begin{figure}[tbp]
   \centering
   \scalebox{0.8}{
   \begin{tabular}{cl}
 \hspace{-.45em}   \includegraphics[height=.25\textwidth, trim={0 0 0 0}, clip]{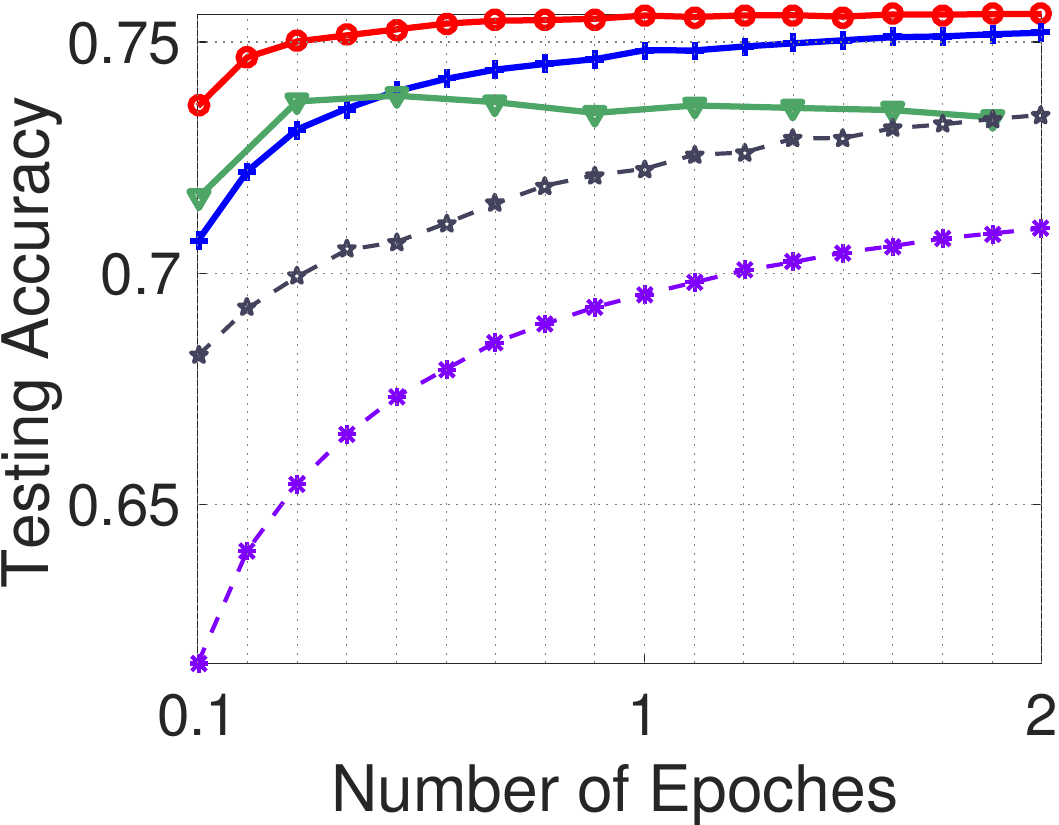}  &   
  \hspace{-.3em}      \includegraphics[height=.25\textwidth, trim={0cm 0 0 0}, clip]{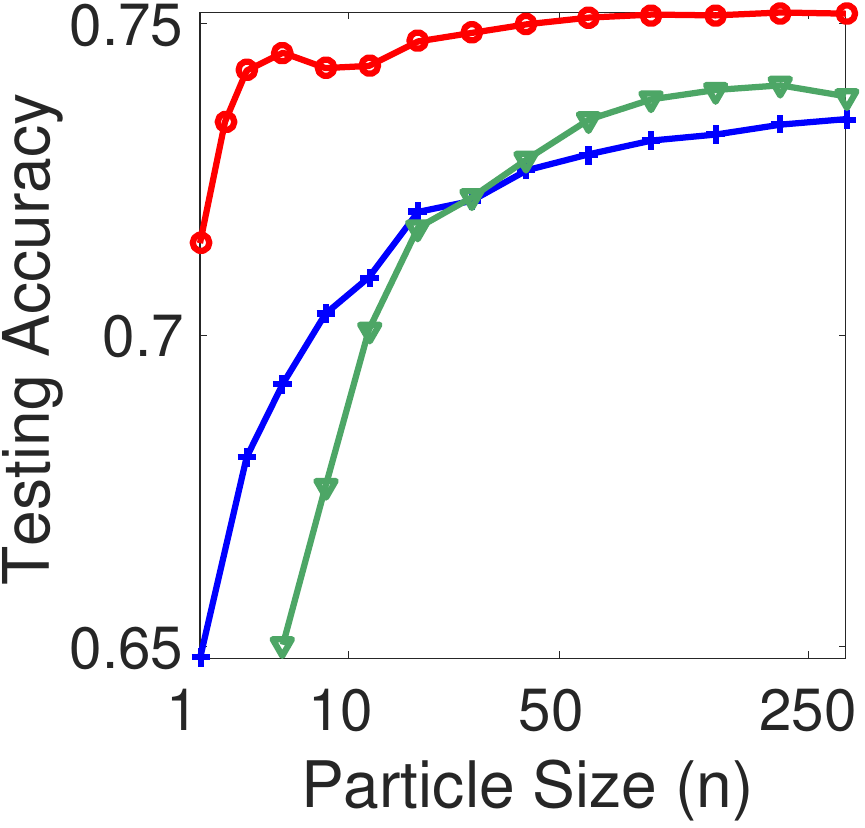} 
      \hspace{0em}\raisebox{6em}{\includegraphics[height=.1\textwidth, trim={0cm 0 0 0}, clip]{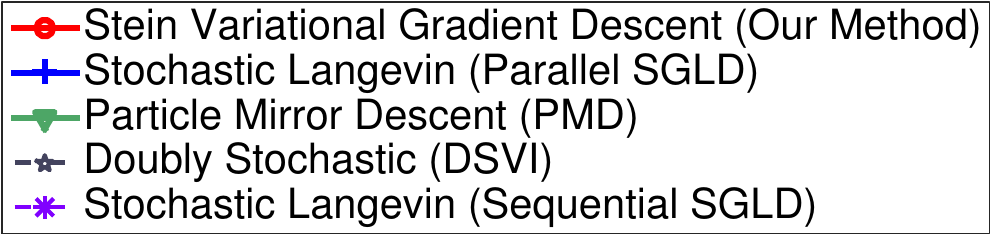}} \\
   {\small   (a) Particle size $n=100$} & {\small (b) Results at 3000 iteration ($\approx$ 0.32 epoches)}
% \hspace{-7.9em}\raisebox{5em}{   \includegraphics[width=.18\textwidth, trim={0 0 0 0}, clip]{figures/leg_LL_bayeslr_covtype}   } \\
%{\small\it (a) Testing Log-likelihood} &  {\small\it(b) Testing Accuracy} & &
   \end{tabular}
   }
   \caption{Results on Bayesian logistic regression on {Covertype} dataset w.r.t. epochs and the particle size $n$. 
   We use $n=100$ particles for our method, parallel SGLD and PMD, and average the last $100$ points for the sequential SGLD. 
   The ``particle-based'' methods (solid lines) in principle require 100 times of likelihood evaluations compare with DVSI and sequential SGLD (dash lines) per iteration,  
   but are implemented efficiently using Matlab matrix operation (e.g., each iteration of parallel SGLD is about 3 times slower than sequential SGLD). 
      %they can be significantly speeded up  using MATLAB matrix operation  
%   The testing loglikelihood (LL) and testing accuracy of different algorithms 
%  w.r.t. running time and number $n$ of particles (
%  DSVI is not a particle based algorithm, and is not shown in the two figures on the right. 
  We partition the data into $80\%$ for training and $20\%$ for testing and average on 50 random trials.  A mini-batch size of $50$ is used for all the algorithms. 
  } 
   \label{fig:covtype}
\end{figure}

We further test the binary Covertype dataset\footnote{\url{https://www.csie.ntu.edu.tw/~cjlin/libsvmtools/datasets/binary.html}}
%~\citep{Bache+Lichman:2013, collobert2002parallel} 
with 
581,012 data points and 54 features. This dataset is too large, and a stochastic gradient descent is needed for speed.
Because NUTS and NPV do not have mini-batch option in their code, we instead compare with
the stochastic gradient Langevin dynamics (SGLD) by \citet{welling2011bayesian},    
the particle mirror descent (PMD) by \citet{dai2016provable}, 
and the doubly stochastic variational inference (DSVI) by \citet{titsias2014doubly}.\footnote{code: \url{http://www.aueb.gr/users/mtitsias/code/dsvi_matlabv1.zip}.}
We also compare with a parallel version of SGLD that runs $n$ parallel chains and take the last point of each chain as the result. 
This parallel SGLD is similar with our method and we use 
the same step-size of $\epsilon_\ell = a/(t+1)^{.55}$ for both as suggested by \citet{welling2011bayesian} for fair comparison; \footnote{We scale the gradient of SGLD by a factor of $1/n$ to make it match with the scale of our gradient in \eqref{equ:updatex}.} 
we select $a$ using a validation set within the training set. 
For PMD, we use a step size of $\frac{a}{N} / (100 + \sqrt{t})$, and RBF kernel $k(x,x')=\exp(-||x-x'||^2/h)$ with bandwidth $h={0.002\times\mathrm{med}^2}$ which is based on the guidance of \citet{dai2016provable} which we find works most efficiently for PMD. 
Figure~\ref{fig:covtype}(a)-(b) shows the results when we initialize our method and both versions of SGLD using the prior $p_0(\alpha) p_0( w | \alpha)$; 
we find that 
PMD tends to be unstable %when initialized with large weights, 
 with this initialization because it generates weights $w$ with large magnitudes, so we divided the initialized weights by 10 for PMD; 
as shown in Figure~\ref{fig:covtype}(a), this gives some advantage to PMD in the initial stage. 
We find our method generally performs the best, followed with the 
parallel SGLD, which is much better than its sequential counterpart; this comparison is of course in favor of parallel SGLD, 
since each iteration of it requires $n=100$ times of likelihood evaluations compared with sequential SGLD.  
However, by leveraging the matrix operation in MATLAB, we find that each iteration of parallel SGLD is only 3 times more expensive than sequential SGLD. 
%Given that the further speedup may be obtained by more efficient parallel computation, 
%We therefore advocate the use of parallel SGLD over the sequential version.  
%The DVSI is a parametric variational method that approximates the posterior with a multivariate Gaussian, and it is not surprising that it is less competitive than the three particle-based methods. 
%Although DVSI only only requires one likelihood evaluation like sequential SGLD, we find it slower than the other algorithms because it needs update the full Gaussian covariance matrix. % of the variational distribution. 
%Given that further speedup can be 
%only costs 4 times 
% parallel SGLD 
%the particle based methods is 
%From Figure~\ref{fig:covtype}(a)-(b)
%(it seems that PMD needs a smaller bandwidth than our method to work  efficiently). 
%which we find get the best performance. 
 %with $a = 5\times10^{-3}$ and $r=0.55$). 
%The typical sequential GLD was also tested, and was found to perform worse than the parallel version. 
\todo{Mirror Descent:   stepsize = $ \frac{1}{N} * 1e4 / (100 + \sqrt{iter})$, where $N$ is the number of observations. kernel, bandwidth =  $\sqrt{0.01*med^2 / \log{N}}$}
\todo{Figure. (a) Our method and Langevin parallel,  $1000:1000:18000$, (iterations); Langevin sequential, $10000:5000:60000$ (iterations); doubly stochastic, $3000:3000:18000$;  PMD,  $1000:2000:17000$ }
\todo{Figure. (b) vary-M(\#particles), fix iterations = 3000}
%
%
%The result is shown in Figure~\ref{fig:covtype}, where we find our method performs the best. 
%We find that the parallel SGLD performs much better than the sequential SGLD
% where we find that our method generally performs the best. 
%outperforms both GLD and DSVI. 
%It is not surprising to see that DSVI is much worse, since it uses a parametric Gaussian distribution to approximate the posterior distribution; DSVI is slower because it needs to update the Gaussian covariance matrix. 
%We find that 
%and the doubly stochastic variational inference by \citet{titsias2014doubly} (NUTS and NPV does not provide mini-batch version in their code). 
%and compare with the stochastic Langevin Dynamics \citep{welling2011bayesian}, and the doubly stochastic variational inference by \citet{titsias2014doubly} (NUTS and NPV does not provide mini-batch version in their code). 
%   
%
\todo{$a = 1; b = 0.01. Stepsize = 5e-3 * (iter + 1)^{-0.55}$}
\todo{mini-batch size = 50}
%\red{Number of particles = 100}
\todo{Initilization, from the prior distribution, except the GMM gradient-free}
%\red{random partition, $80\%$ for training and $20\%$ for testing. Average over 50 trials}
%
\todo{We compared with Stochastic Langevin and DSVI (doubly stochastic variational inference): Langevin, stepsize = $ 5e-5 * (iter + 1)^{-0.55}$ (strategy proposed in their paper). M (number of particles) chains, only the last sample of each chain is used for evaluation. DSVI, default settings in the code.}

%\subsection{Bayesian Neural Network}
\paragraph{Bayesian Neural Network}
We compare our algorithm with the probabilistic back-propagation (PBP) algorithm by \citet{hernandez2015probabilistic} 
on Bayesian neural networks. 
%We take 
%We follow the setting of \citet{XXX}. 
Our experiment settings are almost identity, %the same as that of \citet{hernandez2015probabilistic},
except that we use a $\mathrm{Gamma}(1, 0.1)$ prior for the inverse covariances 
and do not use the trick of scaling the input of the output layer. 
%except that we did not use the trick of scaling the input of each layer 
We use neural networks with one hidden layers, and take 50 hidden units for most datasets, except that we take 100 units for Protein and Year 
which are relatively large; % and we take 100 units; 
all the datasets are randomly partitioned into $90\%$ for training and $10\%$ for testing, and the results are averaged over {$20$} random trials, except for
Protein and Year on which 5 and 1 trials are repeated, respectively. 
We use $\mathrm{RELU}(x)=\max(0, x)$ as the active function, 
whose weak derivative is $\ind[x > 0]$ (Stein's identity also holds for weak derivatives; see
\citet{stein2004use}). 
PBP is repeated using the default setting of the authors' code\footnote{https://github.com/HIPS/Probabilistic-Backpropagation}. 
For our algorithm, we only use $20$ particles, and use AdaGrad with momentum as what is standard in deep learning. 
The mini-batch size is 100 except for Year on which we use 1000. 

We find our algorithm consistently improves over PBP both in terms of the accuracy and speed (except on Yacht); this is encouraging since PBP were specifically designed for Bayesian neural network. We also find that our results are comparable with the more recent results reported on the same datasets \citep[e.g.,][]{li2015stochastic, li2016variational, gal2015dropout} which leverage some advanced techniques that we can also benefit from.  
 \todo{????except we did not use the trick of scale the input of each layer. Standard one hidden layer neural network, 50 hidden units for small dataset, 100 hidden units for Protein and Year. $90\%$ for training and $10\%$ for testing. Nolinear function is relu.
\red{ The settings are almost the same (except we don't use the trick to scale the input of each layer. e.g. input-of-next-layer = output-of-previous-layer / $\sqrt{previous-layer-hidden-units}$). Standard one hidden layer neural network, 50 hidden units for small dataset, 100 hidden units for Protein and Year. $90\%$ for training and $10\%$ for testing. Nolinear function is relu. }
}
\todo{Stepsize. adagrad with momentum. master-stepsize = 1e-3, $historicalgrad = autocorr*historicalgrad + (1-autocorr)*g^2$, where $autocorr=0.9$. batch size = 100 (default). For year, batch size = 1000.}
\todo{Number of particles = 20}
\todo{Hyper-parameters, a = 1, b = 0.1 (gamma distribution)}
\todo{The number of splits is 20 except for the large datasets, which is 5 and 1 for Protein and Year, respectively. (same settings in pbp paper)}
\todo{$bandwidth = \sqrt{0.5* meandistance / log(M+1)}$ (M is the number of particles)}

\begin{table}[H]
\centering{\small
\scalebox{0.95}{
\begin{tabular}{l|cc|cc|cc}
\hline
& \multicolumn{2} {c} {Avg. Test RMSE} & \multicolumn{2} {|c}{Avg. Test LL} & \multicolumn{2}{|c}{Avg. Time (Secs)} \\
\textbf{Dataset} & {\bfseries PBP} & {\bfseries Our Method} & {\bfseries PBP}& {\bfseries Our Method} & {\bfseries PBP} & {\bfseries Ours} \\
\hline
Boston& $ 2.977 \pm 0.093$ & $ \pmb{2.957 \pm 0.099}$ & $-2.579\pm 0.052$ & $ \pmb{-2.504 \pm 0.029}$ & $18$ & $ \pmb{16}$\\
Concrete& $5.506 \pm 0.103$ & $\pmb{5.324\pm 0.104}$& $-3.137\pm 0.021$& $\pmb{-3.082 \pm 0.018}$& $ 33 $ & $ \pmb{24} $\\
Energy& $1.734 \pm 0.051$ & $\pmb{1.374\pm 0.045}$ & $-1.981 \pm 0.028$ & $\pmb{-1.767 \pm 0.024}$ & $25$ & $\pmb{21}$\\
Kin8nm& $0.098\pm 0.001$ & $\pmb{0.090 \pm 0.001}$ & $ ~~~0.901 \pm 0.010$ & $\pmb{~~~0.984 \pm 0.008}$ & $118$& $\pmb{41}$\\
Naval& $0.006\pm 0.000$ & $\pmb{0.004\pm 0.000}$ & $~~~3.735 \pm 0.004$ & $\pmb{~~~4.089\pm 0.012}$ & $173$ & $\pmb{49}$ \\
Combined& $4.052 \pm 0.031$ & $\pmb{4.033 \pm 0.033}$& $-2.819 \pm 0.008$ & $\pmb{-2.815 \pm 0.008}$ & $136$ & $\pmb{51}$ \\
Protein& $4.623\pm 0.009$ & $\pmb{4.606 \pm 0.013}$ & $-2.950 \pm 0.002$ & $\pmb{-2.947 \pm 0.003}$ &$682$ & $\pmb{68}$ \\
Wine& $0.614\pm 0.008$ & $\pmb{0.609\pm 0.010}$ & $-0.931 \pm 0.014$& $\pmb{-0.925 \pm 0.014}$ & $26$ & $\pmb{22}$\\
Yacht&$ \pmb{0.778\pm 0.042}$ & $0.864\pm 0.052$& $ \pmb{-1.211\pm 0.044}$ & $-1.225\pm 0.042$ & $25$ & $25$ \\
Year& $8.733\pm \mathrm{NA}~~~$ & $\pmb{8.684 \pm \mathrm{NA}}~~~$& $-3.586 \pm  \mathrm{NA}~~~$ & $\pmb{-3.580 \pm \mathrm{NA}~~~}$ & $7777$& $\pmb{684}$\\
%Boston& $3.067 \pm 0.098$ &  $ \mathbf{2.925\pm 0.093^*}$ & $-2.566\pm 0.046$ & $\mathbf{ -2.487\pm 0.029^*}$ & $\mathbf{10}$ & $15$\\
%Concrete& $5.616\pm 0.107$ & $\mathbf{5.398\pm 0.118^*}$& $-3.151\pm 0.020$& $\mathbf{-3.097\pm 0.021^*}$& $\mathbf{17}$ & $20$\\
%Energy& $1.858\pm 0.046$ & $\mathbf{1.411\pm 0.047^*}$ & $-2.048\pm 0.024$ & $\mathbf{-1.762\pm 0.033^*}$ & $14$ & $14$\\
%Kin8nm& $0.098\pm 0.001$ & $\mathbf{0.094\pm 0.001}^*$ & $~~~0.899\pm 0.008$ & $\mathbf{~~~0.937\pm 0.009^*}$ & $112$& $\mathbf{ 34}$\\
%Naval& $0.006\pm 0.000$ & $\mathbf{0.005\pm 0.000^*}$ & $~~~3.735\pm 0.005$ & $\mathbf{~~~3.789\pm 0.006^*}$ & $166$ & $\mathbf{ 43}$ \\
%Combined& $4.046\pm 0.031$ & $\mathbf{4.041\pm 0.032}$& $-2.818\pm 0.007$ & $\mathbf{-2.815 \pm 0.008}$ & $128$ & $\mathbf{ 41}$ \\
%Protein& $4.627\pm 0.007$ & $\mathbf{4.626\pm 0.009}$ & $-2.951\pm 0.001$ & $\mathbf{-2.950\pm 0.002}$ &$653$ & $\mathbf{ 50}$ \\
%Wine& $0.614\pm 0.008$ & $\mathbf{0.611\pm 0.009}$ & $-0.932\pm 0.014$& $\mathbf{-0.921\pm 0.015^*}$ & $25$ & $\mathbf{ 15}$\\
%Yacht&$1.024\pm 0.042$ & $\mathbf{0.912\pm 0.051^*}$& $-1.632\pm 0.014$ & $\mathbf{-1.357\pm 0.043^*}$ & $\mathbf{7}$ & $19$ \\
%Year& $8.835\pm \mathrm{NA}$ & $\mathbf{8.825\pm \mathrm{{NA}}}$& $-3.598\pm  \mathrm{NA}~~~$ & $\mathbf{-3.596\pm \mathrm{{NA}}~~~}$ & $10140$& $\mathbf{688}$\\
\hline
\end{tabular}
}
}
%\raisebox{0em}{``*'' denotes the performance of the two algorithms are significantly different by {\em t-test} (0.05 level).}% is applied to determine if the performances of two algorithms are significantly different.}
\end{table}

%\vspace{-3\baselineskip}
\section{Conclusion}
\label{sec:conclusion}

We propose a simple general purpose variational inference algorithm for fast and scalable Bayesian inference. 
%and demonstrate its efficiency using empirical results.  In some sense, we believe algorithm provides a \emph{natural counterpart} of gradient descent for Bayesian inference. \todo{(for a while, we believe it was Langevin Daynamics, but it actually adds noise , and relates to the fundamental drawback of MCMC)}
%and demonstrate its efficiency in practice. 
%Our main theoretical findings reveals the connection between KL divergence and kernelized Stein discrepancy, which has its own independent interest. 
%This work leaves a lot of open directions that we would like to address in the future, 
Future directions include more theoretical understanding on our method, 
more practical applications in deep learning models, and other potential applications of our basic Theorem in Section~\ref{sec:steinKL}. 
%We will also make our code publicly available. 
%our main theoretical findings reveals the connection between KL divergence and kernelized Stein discrepancy, which has its own independent interest. 
%We would like to 
%\newpage
%\renewcommand{\bibsection}{\subsubsection*{References}}
%\small
%\bibliographystyle{myunsrtnat}
%\bibliography{bibrkhs_stein}

%{\small % \setstretch{-5} %\renewcommand{\baselinestretch}{.1} 
%\begin{spacing}{-1}
%\begingroup%\sin­glespac­ing
%\setstretch{-1}
%\bibliographystyle{myunsrtnat}
%\bibliography{bibrkhs_stein}%}
%\end{spacing}
%}
%\endgroup
%}
%\clearpage \newpage
%\appendix 
%
%\end{document}

%\newpage\clearpage
%
{%\small % \setstretch{-5} %\renewcommand{\baselinestretch}{.1} 
%\begin{spacing}{-1}
%\begingroup%\sin­glespac­ing
%\setstretch{-1}
\bibliographystyle{myunsrtnat}
\bibliography{bibrkhs_stein}%}
%\end{spacing}
%}
%\endgroup
}
%\clearpage \newpage
%\appendix 

\newpage\clearpage
\appendix
\numberwithin{equation}{section}

\section{Proof of Theorem~\ref{thm:dkl}}

\begin{lem}\label{lem:complex}
Let $q$ and $p$ be two smooth densities, and $\T = \T_\epsilon(x)$ an one-to-one transform on $\X$ indexed by parameter $\para$, and $\T$ is differentiable w.r.t. both $x$ and $\epsilon$. 
Define $q_{[\T]}$ to be the density of $z=\T_\epsilon(x)$ when $x\sim q$, and $\score_p = \nabla_x \log p(x)$, we have
$$\nabla_\para \KL(q_{[\T]} ~|| ~ p )  =  
\E_q \big [ \score_p( \T(x)) ^\top \nabla_{\para}  \T(x)  +  \trace((\nabla_x  \T(x))^{-1}  \cdot \nabla_\para \nabla_{x} \T(x) ) \big ].
$$
% - \E_{x\sim q} [\nabla_{\para} \log p_{[T^{-1}]}(x)]. $$
\end{lem}
\begin{proof}%[Proof of Theorem~\ref{thm:dkl}]
%Define $\tilde T(x) = T^{-1}(x)$ for convenience, 
Denote by $p_{[\T^{-1}]}(z)$ the density of $z = \T^{-1}(x)$ when $ x \sim p(x)$, then 
$$
q_{[\T^{-1}]}(x) = q(\T(x)) \cdot |\det(\nabla_x \T (x))|. 
$$
By the change of variable, we have 
$$
%\KL(q_{[T]} ~|| ~ p ) = \KL(q ~||~ p_{- \para}), 
\KL(q_{[\T]} ~|| ~ p ) = \KL(q ~||~ p_{[\T^{-1}]}), 
$$
and hence 
$$
\nabla_\para \KL(q_{[\T]} ~|| ~ p )  =   - \E_{x\sim q} [\nabla_{\para} \log p_{[\T^{-1}]}(x)]. 
$$
We just need to calculate $\log p_{[\T^{-1}]}(x)$; define $\score_p(x) = \nabla_x \log p(x)$, we get
$$
\nabla_\para \log p_{[\T^{-1}]} (x) =  \score_p( \T(x)) ^\top \nabla_{\para}  \T(x)  +  \trace((\nabla_x  \T(x))^{-1}  \cdot \nabla_\para \nabla_{x}  \T(x) ). 
$$
\end{proof}

\begin{proof}[Proof of Theorem~\ref{thm:dkl}]
When $\T(x) = x + \para \ff(x)$ and $\para = 0$, we have 
\begin{align*}
\T(x) = x, &&
 \nabla_\para \T(x) = \ff(x), &&
 \nabla_x  \T(x) = I, &&
 \nabla_\para \nabla_{x}  \T(x) = \nabla_x \ff(x), 
\end{align*}
where $I$ is the identity matrix. Using Lemma~\ref{lem:complex} gives the result. 
%we have % then gives
%$$
%\nabla_\para \log p_{[T^{-1}]} (x) ~ \big |_{\para = 0}  = \trace(   \score_p(x) \phi(x)^\top  + \nabla_x \phi(x) ) = \trace(\stein_p \phi(x)). 
%$$
%This concludes the proof. 
%Taking the derivative for KL divergence gives
%\begin{align}
%\nabla_{\para}\KL(q_{[T]} ~||~ p ) %~\big|_{\para = 0} 
%& =  \E_{x \sim q_{[T]}}  \big [  \nabla_{\para} \log q_{[T]}(x) (  \log q_{[T]}(x) - \log p(x)  )  \big] ~ + ~ \int_x \nabla_\para q_{[T]}(x ) dx.  \\
%& =  \E_{x \sim q_{[T]}}  \big [  \nabla_{\para} \log q_{[T]}(x) (  \log q_{[T]}(x) - \log p(x)  )  \big]  \\
%& =  \E_{x \sim q_{[T]}}  \big [  (\score_{q_{[T]}}(x) ^\top  \nabla_\para \tilde T(x) )(  \log q_{[T]}(x) - \log p(x)  )   \big]  \\
%& =  \E_{x \sim q_{[T]}}  \big [  (\score_{q_{[T]}}(x) ^\top  \phi(x) )(  \log q_{[T]}(x) - \log p(x)  )   \big]  \\
%& =  - \E_{x \sim q_{[T]}}  \big [  \nabla_{x} (\phi(x) (  \log q_{[T]}(x) - \log p(x) )  )   \big]  \\
%& =   - \E_{x \sim q_{[T]}}  \big [ \nbla_{\para} ]% \nabla_{\para} \log q_{[T]}(x) (  \log q_{[T]}(x) - \log p(x)  )  \big] 
%~ + ~ \int_x \nabla_\para q_{[T]}(x ) dx 
%\E_{x\sim q_{[T]}} [\nab]
%\end{align}
%where the second term equals zero because $ \int_x \nabla_\para q_{[T]}(x ) dx   = \nabla_{\para} \int_x q_{[T]}(x) dx = 0$ (we can exchange the order of $\nabla_\para$ and $\int $ because $\nabla_\para q_{[T]}(x)$ is a continuous function of $x$ and $\para$). 
\end{proof}

\section{Proof of Theorem~\ref{thm:fungrad}}
%\section{Functional Gradient}
%Let $f \in $
Let $\H^d = \H \times \cdots \times \H$ be a vector-valued RKHS, and $F[f]$ be a functional on $f$. The gradient $\nabla_f F[f]$ of $F[\cdot]$ is a function in $\H^d$ that satisfies
$$
F[f + \epsilon g]=  F[f] + \epsilon~  \la \nabla_f F[f], ~  g \ra_{\H^d} + \Od(\epsilon^2). 
$$

\begin{proof}
Define $F[f] =  \KL(q_{[x+f(x)]} ~||~ p) =
 \KL(q ~||~ p_{[(x+f(x))^{-1}]} )$, we have 
\begin{align*}
%F[f] =  
F[f+\epsilon g] 
& = \KL(q ~||~ p_{[(x+f(x)+\epsilon g(x))^{-1}]} )  \\
& = \E_q [ \log q(x) - \log p(x+ f(x) + \epsilon g(x)) - \log \det(I + \nabla_x f(x) + \epsilon \nabla_x g(x))], 
%& = \E_q [ \log q(x) - \log p(f(x) + \epsilon g(x)) - \log \det(f(x) + \epsilon g(x))] \\
%& = \E_q [ \log q(x) - \log p(f(x) + \epsilon g(x)) - \log \det(f(x) + \epsilon g(x))] \\
\end{align*}
and hence we have %can show that 
$$
F(f + \epsilon g) - F[f] =  - \Delta_1 -  \Delta_2,
$$
where 
\begin{align*}
&\Delta_1 =  \E_q [\log p(x+ f(x) + \epsilon g(x)) ] -  \E_q [\log p(x+f(x))] ,  \\
&\Delta_2 = \E_q [ \log \det(I + \nabla_x f(x) + \epsilon \nabla_x g(x))] -  \E_q [\log \det(I+\nabla_x f(x))] . 
\end{align*}
For the terms in the above equation, we have
\begin{align*}
\Delta_1 &  = \E_q [\log p(x + f(x) + \epsilon g(x)) ] -  \E_q [\log p(x + f(x))]  \\
&    =   \epsilon ~ \E_q [\nabla_x \log p(x + f(x)) \cdot  g(x) ] + \Od(\epsilon^2) \\
&    =   \epsilon ~ \E_q [ \nabla_x \log p(x + f(x)) \cdot  \la  k(x, \cdot), ~ g  \ra_{\H^d}] + \Od(\epsilon^2) \\
&    =   \epsilon ~\la  \E_q [ \nabla_x \log p(x + f(x)) \cdot    k(x, \cdot)],  ~ g  \ra_{\H^d} + \Od(\epsilon^2),
\end{align*}
and 
\begin{align*}
\Delta_2& =  \E_q [ \log \det(I + \nabla_x f(x) + \epsilon \nabla_x g(x))] -  \E_q [\log \det(I + \nabla_x f(x))]  \\
&    =   \epsilon ~ \E_q[ \trace( (I + \nabla_x f(x))^{-1}  \cdot \nabla_x g(x) )] ~+~ \Od(\epsilon^2) \\
&    =   \epsilon ~ \E_q[ \trace( (I + \nabla_x f(x))^{-1}  \cdot \la  \nabla_x k(x, \cdot),  ~ g \ra_{\H^d} ] ~+~ \Od(\epsilon^2) \\
&    =   \epsilon ~  \la  \E_q[ \trace( (I + \nabla_x f(x))^{-1}  \cdot \nabla_x k(x, \cdot)], ~  g \ra_{\H^d}  ~+~ \Od(\epsilon^2) \\
\end{align*}
and hence 
$$
F(f + \epsilon g) - F[f] = \epsilon ~ \la \nabla_f F[f], ~ g\ra_{\H^d} ~+~ \Od(\epsilon^2),  
$$
where 
\begin{align}\label{equ:gdf}
 \nabla_f F[f] =  -  \E_q [ \nabla_x \log p(x + f(x)) +  \trace( (I + \nabla_x f(x))^{-1}  \cdot \nabla_x k(x, \cdot) ]. 
\end{align}
Taking $f = 0$ then gives the desirable result. 
\end{proof}
%\bibliographystyle{unsrtnat}
%\bibliography{bibfilename}
%\clearpage \newpage
%\appendix 

%\blue{
%\begin{thm}
%Assume $\T(x) = x + \epsilon \ff(x)$, where $\ff \in \H^d$. Let kernel $k(x,x')$ be the kernel of $\H$ and assume $k(x, x')$ is continuously differentiable. There exist a $\epsilon_0$, such that $\T(x)$ is an one-to-one map on $\X$ when $|\epsilon| \leq \epsilon_0$.  
%\end{thm}
%\begin{proof}
%Following \citet{zhou2008derivative},  $\nabla_x \ff(x) \in \H^d$, and hence $|\nabla_x \ff(x)| = | \la \nabla_x \ff, k(x, x') \ra| \leq || \la \nabla_x \ff||_{\H} ||k(x,x')||_{\H}$ ....... 
%\end{proof}
%}

\section{Connection with de Bruijn's identity and Fisher Divergence}
%Our results above have a close connection with Fisher divergence and \emph{ de Bruijn's identity}.  
%Note that if we instead find the deepest descent 
%Fisher divergence between $p$ and $q$ is defined as $\F(q,~p) = \E_q[|| \nabla_x \log p - \nabla_x \log q||^2_2]$. 
%If we search the steepest descent direction within ball $\{  \ff  \colon \E_q[|| \ff ||_2^2] \leq  \F(q,~p) \}$ of space $L_2(\X, q)$, we get $\ff^o(x) =  \log p - \log q$ by Theorem~XXX of \citet{liu2016kernelized}, in which case 
%If we take $\ff^{{\scriptstyle\Delta}}_{pq}(x) = \nabla_x \log p -  \nabla_x \log q$ in \eqref{equ:pdir}, we get 
If we take $\ff_{q,p}(x) = \nabla_x \log p(x) -  \nabla_x \log q(x)$ in \eqref{equ:pdir}, we can show that \eqref{equ:pdir} reduces to %get % (see Appendix) 
$$\nabla_\para \KL(q_{[\T]} ~||~ p) \big |_{\para = 0} = - \F(q, ~ p),$$
where $\F(q, ~ p)$ is the Fisher divergence between $p$ and $q$, defined as 
$$\F(q,~p) = \E_q[|| \nabla_x \log p - \nabla_x \log q||^2_2].$$  
Note that this can be treated as a deterministic version of \emph{de Bruijn's identity} \citep{cover2012elements, lyu2009interpretation}, which draws similar connection between KL and Fisher divergence, but uses randomized linear transform $\T(x) = x+ \sqrt{\epsilon} \cdot \xi$, where $\xi$ is a standard Gaussian noise.  

% with variance $\epsilon$. 
%$\nabla_\para \KL(q_{[\T]} ~||~ p) = $ 

%\newpage
\section{Additional Experiments}
We collect additional experimental results that can not fitted into the main paper. % due to the space constraint. 

\begin{figure}[htbp]
\label{fig:uncer}
   \centering
   \begin{tabular}{cc}
\raisebox{1em}{   \includegraphics[height=.25\textwidth, trim={0cm 0 0 0}, clip]{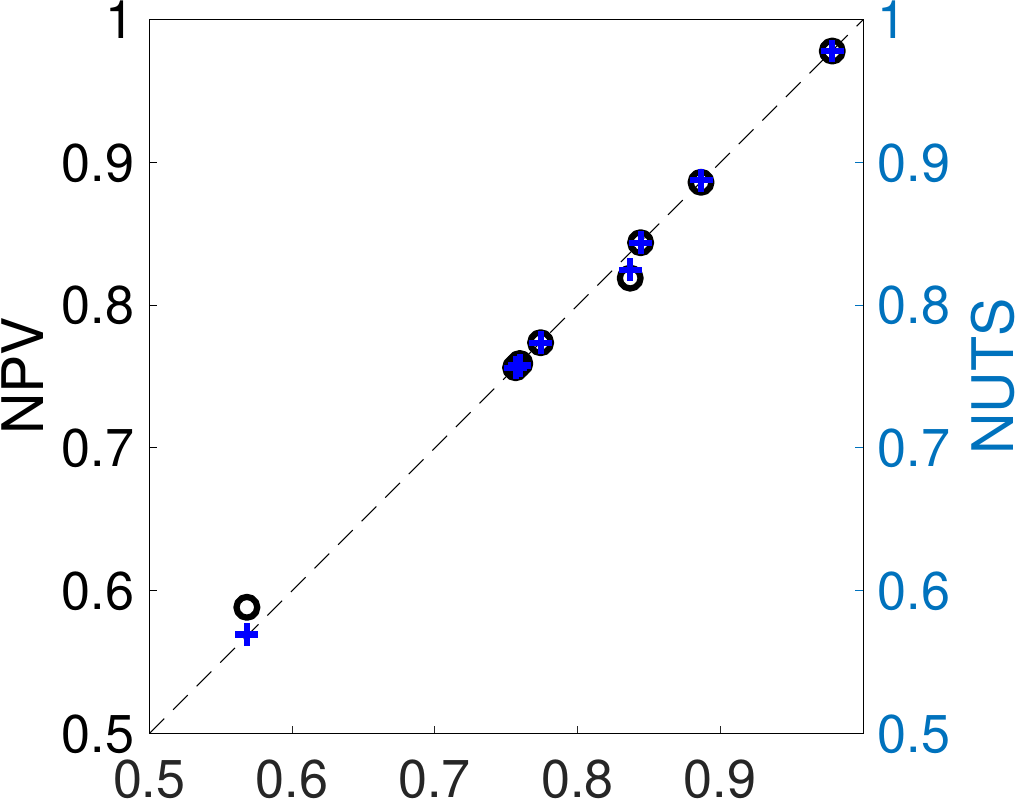}} &
\raisebox{1em}{   \includegraphics[height=.25\textwidth, trim={0cm 0 0 0}, clip]{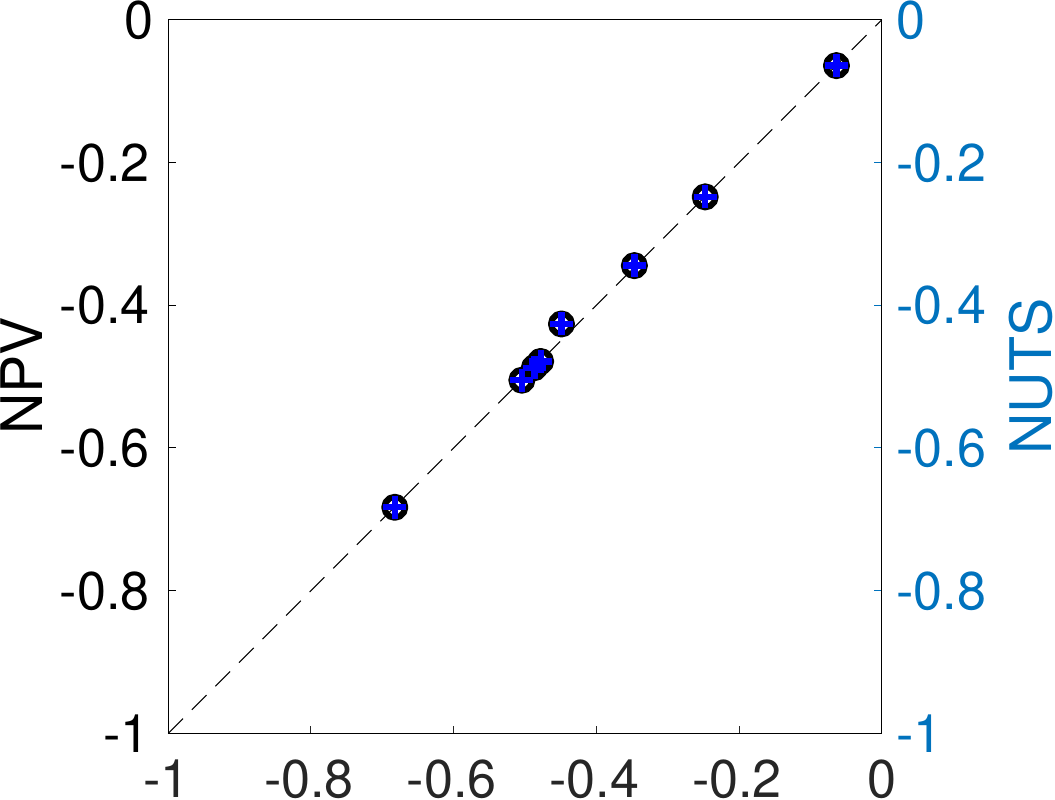} } \\
   {\small (a) Testing Accuracy} & {\small (b) Testing Log-Likelihood}
   \\
   \end{tabular}
   \setlength{\unitlength}{\textwidth}
   \begin{picture}(0,0)
   \put(-.6,-.11){{\scriptsize Our Method}}
   \put(-.25,-.11){{\scriptsize Our Method}}   
   \end{picture}      
   \caption{Bayesian logistic regression on the 8 datasets studied in \citet{gershman2012nonparametric}.
   We find our method performs similarly as NPV and NUTS on all the 8 datasets. 
   }
\end{figure}

\begin{figure}[htbp]
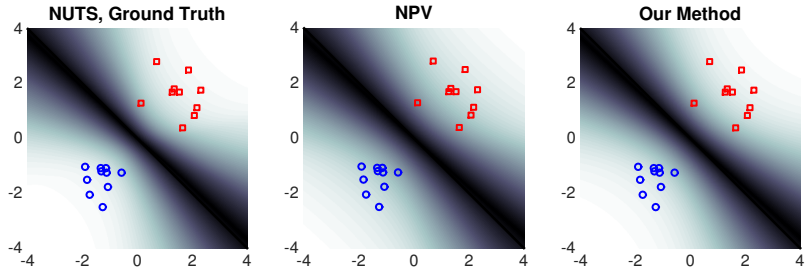

\label{fig:uncer}
   \centering
   \begin{tabular}{cccc}
   \includegraphics[height=.25\textwidth, trim={0cm 0 0 0}, clip]{figures/star_uncertainty} &
   \includegraphics[height=.25\textwidth, trim={0cm 0 0 0}, clip]{figures/npv_uncertainty} &
   \includegraphics[height=.25\textwidth, trim={0cm 0 0 0}, clip]{figures/stein_uncertainty} &
   \\
   \end{tabular}
   \caption{Bayesian logistic regression. The posterior prediction uncertainty as inferred by different approaches on a toy data.}
\end{figure}

\subsection{Bayesian Logistic Regression on Small Datasets}
%\cutspace{
We consider the Bayesian logistic regression model for binary classification, on which the regression weights $w$ is assigned with a Gaussian prior $p_0(w) = \normal(w, \alpha^{-1})$
and $p_0(\alpha) = \Gamma(\alpha, a, b)$, and apply inference on posterior $p(x \cd D)$, where $x = [w, \log \alpha]$. The hyper-parameter is taken to be $a=1$ and $b = 0.01$. 
This setting is the same as that in \citet{gershman2012nonparametric}. 
We compared our algorithm with  the no-U-turn sampler (NUTS)\footnote{code: http://www.cs.princeton.edu/~mdhoffma/} \citep{homan2014no} and 
non-parametric variational inference (NPV)\footnote{code: http://gershmanlab.webfactional.com/pubs/npv.v1.zip} on the 8 datasets ($N>500$) as used in \citet{gershman2012nonparametric}, in which we use {$100$} particles,
NPV uses {100} mixture components, and NUTS uses {1000} draws with {$1000$} burnin period. 
We find that all these three algorithms almost always performs the same across the 8 datasets (See Figure ~ in Appendix), and this is consistent with Figure 2 of \citet{gershman2012nonparametric}. 

We further experimented on a toy dataset with only two features and visualize the prediction probability of the three algorithms in Figure~\ref{fig:uncer}. 
We again find that all the three algorithms tend to perform similarly. Note, however, that NPV is relatively inconvenient to use since it requires the Hessian matrix, and  NUTS tends to be very small when applied on massive datasets. %big data settings. 
%}

\end{document}